\pdfoutput=1
\ifdefined\pdfminorversion\pdfminorversion=7\fi
\documentclass[11pt]{article}
\usepackage[T1]{fontenc}
\usepackage{lmodern}
\usepackage[a4paper,margin=2.5cm]{geometry}
\usepackage{amsmath,amssymb,amsfonts}
\usepackage{graphicx}
\usepackage{booktabs}
\usepackage{array}
\usepackage{adjustbox}
\usepackage{rotating}
\usepackage[super,sort&compress,comma]{natbib}
\usepackage{longtable}
\usepackage{pdflscape}
\usepackage{xcolor}
\usepackage{setspace}
\usepackage{url}
\usepackage[hidelinks]{hyperref}
\usepackage[font=small,labelfont=bf,labelsep=period]{caption}

\newcommand{\pp}{\,\text{pp}}
\newcommand{\arm}[1]{#1}
\newcolumntype{L}[1]{>{\raggedright\arraybackslash\hyphenpenalty=10000\exhyphenpenalty=10000}p{#1}}

\newcommand{\NAGTtableMain}{\begin{sidewaystable}
\centering
\caption{Main cross-dataset and SFDA benchmark: $J=\tfrac12(\mathrm{S\text{-}F1}+\mathrm{V\text{-}F1})$ (\%) with 95\% patient-cluster bootstrap CI. Icentia11k all-comers is the pre-specified confirmatory analysis; all other columns are development or secondary analyses. Final configuration: relA RR head, $\delta_S=1$, $\delta_F=0.5$, $\alpha=1$, $\tau=0.15$, six encoder seeds. Sources: MIT-BIH-44 for Icentia11k, INCART and SVDB; Track A (14-patient) and Track B (DS1-21) for MIT-BIH DS2; fold sources excluding the target patients for DS1-CV.}
\label{tab:main}
\footnotesize
\setlength{\tabcolsep}{2.6pt}
\begin{adjustbox}{max width=\textwidth}
\begin{tabular}{llcccccccc}
\toprule
Method & Group & Icentia all-comers & Icentia non-AF & Icentia test-75 & MIT-BIH DS1-CV & DS2 A & DS2 B & INCART & SVDB \\
\midrule
G0 (no adaptation) & none & 34.5 [27.4, 41.8] & 43.9 [36.5, 50.3] & 41.1 [20.0, 53.8] & 39.6 [25.2, 54.2] & 47.9 [34.3, 60.4] & 44.1 [30.4, 56.5] & 67.9 [46.6, 81.0] & 43.7 [36.3, 51.7] \\
AdaBN & SFDA & 24.7 [19.7, 29.9] & 31.2 [25.5, 36.5] & 22.1 [9.9, 34.5] & 34.2 [21.8, 45.5] & 36.7 [22.5, 47.8] & 31.0 [18.0, 42.7] & 65.9 [46.0, 77.1] & 38.9 [32.1, 46.2] \\
Tent & SFDA & 28.3 [22.8, 33.9] & 36.2 [30.0, 42.0] & 27.7 [14.2, 41.3] & 37.5 [24.4, 50.8] & 39.5 [24.3, 53.5] & 34.6 [20.8, 48.3] & 69.9 [47.6, 80.5] & 40.6 [33.0, 48.8] \\
T3A & SFDA & 25.6 [19.7, 31.3] & 32.3 [26.3, 37.4] & 30.3 [12.0, 42.4] & 44.6 [35.8, 51.0] & 45.7 [34.6, 51.0] & 44.8 [33.5, 51.1] & 52.4 [43.8, 62.5] & 50.7 [44.4, 57.2] \\
RPA-Recenter & SFDA & 42.8 [35.8, 49.0] & 54.8 [49.2, 59.4] & 44.3 [24.2, 57.5] & 50.9 [40.9, 62.0] & 50.3 [37.6, 59.9] & 46.6 [33.4, 55.0] & 67.6 [49.8, 78.7] & 62.8 [55.9, 69.3] \\
GN & geometric & 35.2 [30.0, 39.5] & 37.9 [33.1, 41.5] & 39.8 [23.4, 47.1] & 46.0 [36.7, 55.0] & 52.1 [41.1, 64.7] & 50.2 [39.5, 59.7] & 57.2 [45.5, 69.0] & 44.0 [38.9, 48.8] \\
GA\_All & geometric & 44.7 [37.8, 50.9] & 56.7 [51.1, 61.3] & 47.1 [27.0, 60.1] & 49.1 [37.2, 61.9] & 52.2 [39.2, 61.4] & 47.4 [35.4, 55.4] & 69.3 [50.6, 80.4] & 62.0 [55.2, 68.5] \\
GA\_SV & geometric & 34.9 [28.6, 41.0] & 44.1 [37.9, 49.9] & 36.3 [21.4, 50.9] & 54.1 [39.2, 68.5] & 53.9 [43.2, 62.0] & 54.0 [42.3, 61.3] & 68.7 [47.1, 78.2] & 58.2 [49.8, 66.5] \\
\textbf{GA\_NS (proposed)} & \textbf{proposed} & \textbf{48.7 [42.1, 54.7]} & \textbf{59.8 [53.9, 64.8]} & \textbf{47.1 [30.6, 63.8]} & \textbf{55.6 [41.4, 69.7]} & \textbf{56.0 [46.6, 63.9]} & \textbf{55.6 [45.0, 62.5]} & \textbf{72.7 [51.0, 82.1]} & \textbf{61.3 [53.5, 69.0]} \\
GA\_NS\_Oracle & diagnostic & 50.1 [43.9, 55.7] & 60.8 [55.3, 65.5] & 50.6 [33.7, 65.6] & 55.6 [43.0, 70.1] & 56.3 [46.6, 65.3] & 56.1 [45.6, 63.2] & 74.4 [51.6, 84.2] & 62.2 [54.5, 70.0] \\
\midrule
Records / patients & & 973 / 493 & 919 / 470 & 145 / 73 & 21 / 21 & 22 / 22 & 22 / 22 & 75 / 32 & 78 / 78 \\
\bottomrule
\end{tabular}
\end{adjustbox}
\end{sidewaystable}
}

\newcommand{\NAGTtableAblation}{\begin{table*}[!tp]
\centering
\caption{Stepwise ablation and error flow (counts are 6-seed means). (a) Block A: MIT-BIH development patients (dev7; in-sample hyperparameter selection). (b) Block B1: MIT-BIH DS2 (Track B source). (c) Block B2: INCART (MIT-BIH-44 source). (d) Block C: Icentia11k (pre-specified confirmatory evaluation; final configuration for every arm).}
\label{tab:ablation}
\scriptsize
\setlength{\tabcolsep}{3pt}
\par\smallskip\textbf{(a)} Block A: MIT-BIH dev7\par\smallskip
\begin{adjustbox}{max width=\textwidth}
\begin{tabular}{p{5.2cm}lcccp{4.2cm}}
\toprule
Step & Data / seeds & $J$ (G0) & $J$ (GN) & $J$ (GA) & Note \\
\midrule
A1 Baseline specification: beat-weighted prototypes, $\tau=1$, oracle anchors & seed 0 & 73.52 & 74.10 & 67.27 & RR logits dominate (range 10.76 vs 1.23) \\
A2 + score balance ($\alpha=0.5$, $\tau=0.25$, trace normalization) & seed 0 & 79.31 & 77.73 & 80.02 & $\Delta\text{V-F1}(\text{GA}-\text{GN})=-3.58\pp$ \\
A3 + \textbf{patient-balanced prototypes} & seed 0 & 78.67 & 79.13 & 83.98 & $+2.4$ to $+10.5\pp$ $J$ (GA) across 10/10 encoder snapshots \\
A4 Frozen encoder recipe ($d=8$, 30 epochs, $\epsilon=0.001$, AIRM; $\alpha=1$, $\tau=0.15$), oracle anchors & seed 0 & 78.99 & 74.68 & 85.85 &  \\
A5 + \textbf{label-free U1 anchors} & seeds 0--2 & 68.02 $\pm$ 9.50 & 74.69 & 70.96 $\pm$ 13.52 & oracle 70.33; V collapse (V-Se $62.7 \pm 22.5\%$) \\
A6 + \textbf{V fixed (GA\_SV)}, locked evaluation & seeds 3--5 & 62.37 & 72.11 & GA\_All 65.74 / GA\_SV 84.89 & V-Se $54.3\% \to 97.8\%$ \\
A7 + \textbf{F fixed (GA\_NS)} & dev7 fold 1 & --- & 73.40 & GA\_SV 85.45 / GA\_NS 84.46 & on dev7, GA\_NS $\approx$ GA\_SV \\
\bottomrule
\end{tabular}
\end{adjustbox}
\par\smallskip\textbf{(b)} Block B1: MIT-BIH DS2\par\smallskip
\begin{adjustbox}{max width=\textwidth}
\begin{tabular}{lcccccccccc}
\toprule
step & J & S-Se & S-PPV & V-F1 & N-Se & S$\to$N & V$\to$S & N$\to$V & N$\to$F & N$\to$S \\
\midrule
G0 (orig. 4-dim RR head) & 47.02 & 22.76 & 13.72 & 76.96 & 78.85 & 963 & 84 & 964 & 4,626 & 2,212 \\
GN & 52.97 & 16.27 & 18.99 & 88.46 & 92.48 & 1,176 & 87 & 279 & 1,472 & 1,023 \\
GA\_All & 47.02 & 22.11 & 12.93 & 77.73 & 91.32 & 964 & 360 & 388 & 781 & 2,033 \\
GA\_SV (V fixed) & 52.63 & 28.92 & 17.99 & 83.10 & 90.46 & 991 & 34 & 792 & 655 & 2,073 \\
GA\_NS (V, F fixed) & 55.88 & 29.21 & 18.18 & 89.37 & 89.33 & 991 & 33 & 306 & 1,558 & 2,072 \\
GA\_NS + \textbf{rate-invariant relA, $\delta_S = 1$} & 56.04 & 25.20 & 23.09 & 88.07 & 92.09 & 1,016 & 13 & 375 & 1,231 & 1,314 \\
GA\_NS + relA + \textbf{$\delta_F = 0.5$ (final)} & 55.61 & 25.24 & 22.98 & 87.25 & 92.71 & 1,017 & 13 & 440 & 927 & 1,323 \\
\bottomrule
\end{tabular}
\end{adjustbox}
\par\smallskip\textbf{(c)} Block B2: INCART\par\smallskip
\begin{adjustbox}{max width=\textwidth}
\begin{tabular}{lcccccccccc}
\toprule
step & J & S-Se & S-PPV & V-F1 & N-Se & S$\to$N & V$\to$S & N$\to$V & N$\to$F & N$\to$S \\
\midrule
G0 (orig. 4-dim RR head) & 46.48 & 92.84 & 5.66 & 82.28 & 67.96 & 20 & 1,795 & 3,515 & 13,767 & 23,955 \\
GN & 51.60 & 77.05 & 8.44 & 88.01 & 82.02 & 309 & 1,773 & 1,247 & 9,774 & 12,118 \\
GA\_All & 49.06 & 85.88 & 6.41 & 86.21 & 75.53 & 19 & 1,872 & 1,765 & 10,814 & 18,918 \\
GA\_SV (V fixed) & 47.02 & 81.65 & 6.66 & 81.72 & 73.84 & 20 & 762 & 5,863 & 9,624 & 18,185 \\
GA\_NS (V, F fixed) & 50.40 & 81.61 & 6.68 & 88.45 & 74.78 & 20 & 759 & 2,143 & 12,166 & 18,146 \\
GA\_NS + \textbf{rate-invariant relA, $\delta_S = 1$} & 73.05 & 76.11 & 44.92 & 89.71 & 86.75 & 39 & 290 & 1,384 & 14,396 & 1,269 \\
GA\_NS + relA + \textbf{$\delta_F = 0.5$ (final)} & 72.70 & 76.18 & 44.70 & 89.15 & 88.16 & 40 & 293 & 1,773 & 12,180 & 1,281 \\
\bottomrule
\end{tabular}
\end{adjustbox}
\par\smallskip\textbf{(d)} Block C: Icentia11k\par\smallskip
\begin{adjustbox}{max width=\textwidth}
\begin{tabular}{lcccccccccc}
\toprule
arm & J & S-Se & S-PPV & V-F1 & N-Se & S$\to$N & V$\to$S & N$\to$V & N$\to$F & N$\to$S \\
\midrule
G0 & 34.54 & 61.02 & 16.56 & 44.12 & 73.05 & 3,837 & 4,840 & 43,280 & 836,275 & 81,241 \\
GN & 35.17 & 18.41 & 4.26 & 63.51 & 94.14 & 16,688 & 3,917 & 11,533 & 103,114 & 94,120 \\
GA\_All & 44.67 & 61.12 & 21.55 & 57.51 & 94.94 & 5,290 & 5,174 & 18,500 & 115,418 & 46,316 \\
GA\_SV & 34.90 & 66.89 & 22.65 & 36.00 & 93.84 & 5,352 & 5,008 & 66,719 & 104,992 & 47,861 \\
\textbf{GA\_NS} & 48.66 & 66.94 & 22.61 & 63.55 & 93.87 & 5,448 & 5,051 & 13,918 & 156,679 & 47,968 \\
GA\_NS\_Oracle & 50.07 & 67.51 & 23.36 & 65.47 & 94.83 & 5,634 & 5,168 & 12,241 & 126,138 & 46,086 \\
\bottomrule
\end{tabular}
\end{adjustbox}
\end{table*}
}

\newcommand{\NAGTtableGeom}{\begin{table*}[!t]
\centering
\caption{Direction of transport (record$\times$seed units with $\geq 10$ evaluation beats of the class; count in expected direction / total, median ratio) and U1 anchor purity across four databases.}
\label{tab:geom}
\scriptsize
\setlength{\tabcolsep}{3pt}
\begin{adjustbox}{max width=\textwidth}
\begin{tabular}{L{4.4cm}L{2.4cm}ccccc}
\toprule
database (source) & lead / rate & N: ratio $<$ 1 & S: ratio $<$ 1 & V: ratio $>$ 1 & U1 anchor purity (N) & prefix N fraction \\
\midrule
MIT-BIH dev7, seeds 0--2 (development) & MLII, 360 Hz & 21/21 (100 \%); 0.27--0.38 & 15/15 (100 \%); 0.64--0.81 & 18/18 (100 \%); 1.75--2.14 & 99.86 \% & --- \\
MIT-BIH dev7, seeds 3--5 (locked evaluation) & MLII, 360 Hz & 21/21 (100 \%); 0.34--0.37 & 13/15 (86.7 \%); 0.69--0.82 & 18/18 (100 \%); 1.88--1.92 & --- & --- \\
MIT-BIH DS2 (Track A source) & MLII, 360 Hz & 132/132 (100.0 \%); 0.249 & 36/48 (75.0 \%); 0.755 & 60/60 (100.0 \%); 1.731 & 96.50--96.85 \% & 89.4 \% \\
MIT-BIH DS2 (Track B source) & MLII, 360 Hz & 132/132 (100.0 \%); 0.290 & 35/48 (72.9 \%); 0.683 & 60/60 (100.0 \%); 1.641 & 96.47--96.81 \% & 89.4 \% \\
INCART (MIT-BIH-44 source) & lead II, 257 Hz & 418/450 (92.9 \%); 0.473 & 47/72 (65.3 \%); 0.879 & 296/318 (93.1 \%); 1.538 & 99.92 \% & 87.9 \% \\
SVDB (MIT-BIH-44 source) & ECG1, 128 Hz & 450/468 (96.2 \%); 0.237 & 349/384 (90.9 \%); 0.590 & 295/312 (94.6 \%); 1.394 & 99.82 \% & 88.6 \% \\
INCART (DS1-21 source) & lead II, 257 Hz & 418/450 (92.9 \%); 0.406 & 52/72 (72.2 \%); 0.885 & 295/318 (92.8 \%); 1.536 & 99.94 \% & 87.9 \% \\
SVDB (DS1-21 source) & ECG1, 128 Hz & 451/468 (96.4 \%); 0.236 & 356/384 (92.7 \%); 0.601 & 297/312 (95.2 \%); 1.418 & 99.86 \% & 88.6 \% \\
\textbf{Icentia11k (MIT-BIH-44 source), U1 anchors} & modified lead I patch, 250 Hz & 5,496/5,838 (94.1 \%); 0.290 & 868/1,026 (84.6 \%); 0.560 & 823/1,092 (75.4 \%); 1.281 & 99.82 \% & 98.6 \% \\
\textbf{Icentia11k (MIT-BIH-44 source), oracle anchors} & modified lead I patch, 250 Hz & 5,538/5,838 (94.9 \%); 0.271 & 887/1,026 (86.5 \%); 0.534 & 802/1,092 (73.4 \%); 1.252 & (all prefix N) & --- \\
\bottomrule
\end{tabular}
\end{adjustbox}
\end{table*}
}

\newcommand{\NAGTtableRuntime}{\begin{table*}[!t]
\centering
\caption{Adaptation cost per record (5-minute prefix; Icentia11k, 104 records, seed 0; median 293 prefix beats and 4236 beats per record). NVIDIA GeForce RTX 5090 for network operations, NumPy (8 threads) for SPD operations. Shared steps (frozen encoding, scoring) are excluded.}
\label{tab:runtime}
\scriptsize
\setlength{\tabcolsep}{3pt}
\begin{adjustbox}{max width=\textwidth}
\begin{tabular}{L{2.9cm}L{4.6cm}cL{3.4cm}cc}
\toprule
method & adaptation step & gradient / backprop & parameters changed per record & re-encode all beats & adaptation time per record, ms (median [IQR]) \\
\midrule
\textbf{GA\_NS} & U1 anchor selection + AIRM Karcher mean of anchors + closed-form 8$\times$8 SPD congruence of P\_N, P\_S & no & none in the network; one 8$\times$8 SPD map (36 free values) & no (frozen features reused) & 5.77 [4.26--6.80] \\
RPA-Recenter & AIRM mean of all prefix beats + 8$\times$8 SPD map applied to 4 prototypes & no & none; one 8$\times$8 SPD map & no & 8.63 [7.28--11.32] \\
T3A (M = 5) & pseudo-label + entropy filter + LE means of 4 prototypes & no & 4 prototypes (4 $\times$ 36 values) & no & 0.23 [0.19--0.25] \\
AdaBN & one forward pass of prefix crops to re-estimate BN statistics & no & 160 BN running statistics (80 channels) & yes & 3.21 [3.07--3.31] \\
Tent (lr 1e-3, 5 epochs) & 25 Adam steps (batch 64) on entropy of the fused score + BN re-estimation & yes & 160 BN affine parameters + 160 BN statistics & yes & 50.58 [42.19--58.81] \\
(re-encoding of all beats, AdaBN/Tent only) & encoder forward + covariance + log map for the whole record & --- & --- & --- & 34.91 [29.30--39.33] \\
\bottomrule
\end{tabular}
\end{adjustbox}
\end{table*}
}

\title{Transport What Shares the Conduction Pathway: Normal-Anchored SPD Geometric Transport for Source-Free Patient Adaptation of Holter and Patch ECG}
\author{JinKook Kim$^{1,}$\thanks{Corresponding author. E-mail: jinkook@huinno.com; jkook1218@gmail.com}\\[3pt]
\normalsize $^{1}$AI Division, HUINNO Co., Ltd., Seoul, Republic of Korea}
\date{}

\begin{document}
\maketitle
\onehalfspacing

\begin{abstract}
\noindent Continuous electrocardiogram (ECG) streams are dominated by normal beats (87.9--98.6\% of the label-free adaptation prefix in our four databases), and patient-specific waveform distortion is not shared by every beat class; under this combination, source-free and test-time adaptation methods that assume class-balanced target batches collapse. We propose Normal-Anchored Geometric Transport (NAGT, \arm{GA\_NS}), which estimates a closed-form symmetric positive-definite congruence map from label-free normal anchors in the first 5\,min of each recording and applies it only to the normal and supraventricular prototypes, keeping the ventricular and fusion prototypes fixed. Decision rules were set on MIT-BIH and the rate-invariant RR head was adopted after the development databases (INCART, SVDB) had been opened; the frozen model was then evaluated once under a pre-specified locked evaluation protocol on the held-out Icentia11k chest-patch cohort. In 493 patients (973 recordings, 3,615,388 evaluated beats), NAGT raised the arrhythmia macro-F1 $J$ from $34.5\%$ to $48.7\%$ ($\Delta J=+14.1$ [+8.6, +19.3]\pp; non-AF $43.9\%\to59.8\%$) and outperformed all five Holm-adjusted comparators, including AdaBN, Tent and T3A. In post hoc analyses of the frozen outputs, ventricular false alarms fell by 68\% (41.1 to 13.2 per hour) and 98.3\% of normal beats stayed outside the alarm classes. Adaptation takes 5.8\,ms per record without back-propagation.
\end{abstract}

\section{Introduction}
\label{sec:intro}

Test-time adaptation of a frozen classifier to each new recording is attractive for long-term physiological monitoring, where labeled target data are unavailable and the source data cannot travel with the device. Continuous ECG streams, however, depart in two ways from the conditions under which source-free domain adaptation (SFDA) and test-time adaptation (TTA) were developed. First, the target class mixture is extremely skewed: normal (N) beats make up 87.9--98.6\% of the label-free adaptation prefix in the four databases studied here, whereas the rare arrhythmia classes are the clinical targets, so the target distribution carries a severe label shift. Second, the covariate shift is class-dependent: the patient- and lead-specific distortion of the QRS waveform is shared by the beats that use the normal conduction pathway but not by ventricular ectopic beats, so the shift acts on a subspace shared by some classes only (a subspace-shared conditional shift) rather than on all classes alike. Inter-patient ECG beat classification exposes both problems and remains the weakest link in automated Holter and wearable-patch analysis pipelines\cite{luz2016survey}. When a model is evaluated on new patients that it did not observe during training (e.g., the MIT-BIH DS1/DS2 inter-patient split protocol\cite{dechazal2004automatic}), the high accuracy achieved in intra-patient splits is lost abruptly. A substantial part of this loss is morphological: even for the same ventricular electrical activation, the QRS waveform projected onto a surface lead varies considerably with the electrode position, the thoracic geometry and the individual myocardial conduction properties. In particular, single-channel wearable chest patches, which are now spreading in clinical and everyday monitoring, have an anatomical lead configuration (e.g., modified lead I) entirely different from the limb leads used in most annotated databases, and therefore further aggravate cross-device and cross-lead domain shift.

Patient-adaptive ECG classifiers have long addressed inter-patient variation with labeled data from each new patient: a mixture of experts with a patient-specific expert trained on the first annotated beats of the recording\cite{hu1997patient}, local classifiers trained on a small number of expert-labeled beats\cite{dechazal2006patient,llamedo2012automatic}, and patient-specific neural networks trained on the first minutes of annotated beats\cite{ince2009generic,kiranyaz2016realtime}. Deep networks trained on large ambulatory datasets\cite{hannun2019cardiologist} narrow but do not remove the inter-patient gap. These approaches require expert labels for every new patient, which source-free adaptation seeks to avoid.

Source-free domain adaptation (SFDA) and test-time adaptation (TTA) have been proposed to mitigate such domain shift without access to the original training data. Representative examples are \arm{AdaBN}\cite{li2018adabn}, which replaces batch-normalization (BN) statistics with those of the target segment; \arm{Tent}\cite{wang2021tent}, which additionally updates the BN affine parameters by back-propagation so as to minimize the Shannon entropy of the predictions; and \arm{T3A}\cite{iwasawa2021t3a}, which adjusts class templates with confident low-entropy pseudo-labels; later work adapts BN statistics to corrupted inputs\cite{schneider2020covariate}, transfers the source hypothesis by information maximization\cite{liang2020shot} and stabilizes TTA on continual, temporally correlated or otherwise non-i.i.d.\ streams\cite{wang2022cotta,gong2022note,niu2023sar,yuan2023rotta}. A separate line treats pure label shift, re-estimating the target class prior by expectation--maximization\cite{saerens2002adjusting} or by confusion-matrix inversion and calibration\cite{lipton2018bbse,alexandari2020bcts}, or regularizing the output posteriors with a Laplacian over the target batch (LAME\cite{boudiaf2022lame}). BN- and entropy-based methods rely on the implicit premise that the target adaptation batch resembles the class-mixture distribution of source training, whereas label-shift corrections assume that the class-conditional distributions are unchanged and only the priors differ. Real continuous patient ECG recordings violate both premises: across the four databases of this study, the proportion of normal (N) beats among the label-free prefix beats reached 89.4--91.3\% in MIT-BIH, 87.9\% in INCART, 88.6\% in SVDB and 98.6\% in Icentia11k, whereas the source encoder was trained with class-balanced batches (25\% per class) for rare-arrhythmia detection. We show that in this setting, among the eight main evaluation sets, \arm{AdaBN} falls below the non-adapted model (\arm{G0}) on all eight, \arm{Tent} on seven and \arm{T3A} on four, and that on Icentia11k all three methods collapse well below \arm{G0}; output-only label-shift corrections (Prior-EM and LAME), applied post hoc to the frozen outputs, behaved inconsistently across databases and stayed well below the proposed method on the confirmatory Icentia11k cohort (Section~\ref{sec:sfda}).

The Riemannian geometry of symmetric positive-definite (SPD) covariance matrices underlies classification and transfer in brain--computer interfaces\cite{barachant2012multiclass,congedo2017primer}, SPD neural networks and their batch normalization\cite{huang2017spdnet,brooks2019riemannian,kobler2022spdbn}, parallel transport between domains\cite{yair2019parallel} and optimal-transport domain adaptation\cite{courty2017optimal}. Riemannian alignment offers a geometric alternative that does not modify network weights directly: it removes session- and subject-specific affine distortions by re-centering covariance matrices to a common reference point\cite{zanini2018transfer,rodrigues2019rpa}. However, if this approach is applied unchanged to ECG beat classification and a single global map moves all class prototypes at once (\arm{GA\_All} and \arm{RPA-Recenter}), every beat class is treated as if it shared the same patient-specific morphological distortion. Ventricular ectopic beats (V) do not: V beats originate from an ectopic focus in the ventricular myocardium and spread by cell-to-cell propagation rather than through the His--Purkinje conduction system. We demonstrate that the map estimated from a patient's normal beats moves the ventricular prototype away from that patient's actual ventricular beats in 75.4--100\% of the record\,$\times$\,seed units across the four databases (Section~\ref{sec:geom}). Conversely, moving only the normal prototype (\arm{GN}) absorbs the supraventricular beats (S), whose waveforms resemble normal beats, into the normal cluster, and moving the fusion prototype (F) as well (\arm{GA\_SV}) causes a balloon effect in which borderline normal beats flow massively into V (Section~\ref{sec:ablation}).

\begin{figure*}[!t]
\centering
\includegraphics[width=\textwidth]{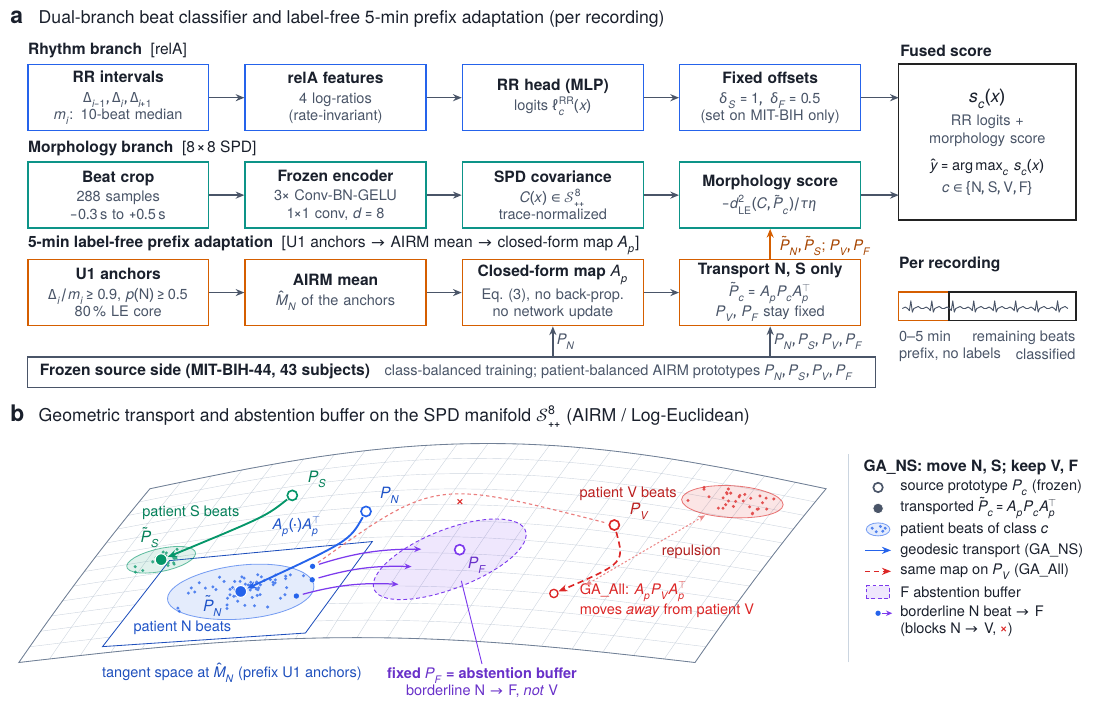}
\caption{Overview of NAGT (\arm{GA\_NS}). (a) Dual-branch model and label-free adaptation pipeline. Each beat is encoded by a frozen convolutional encoder into an $8\times8$ SPD covariance descriptor $C(x)$ and scored by its Log-Euclidean distance to the class prototypes; a parallel rate-invariant relative-RR branch (relA) adds RR logits and the fixed offsets $\delta_S=1$, $\delta_F=0.5$. From the label-free 5-min prefix, U1 normal anchors (non-premature beats with $p(\mathrm{N})\geq0.5$ under the four-dimensional RR head, 80\% core of the patient's own covariance cloud) define the AIRM mean $\hat{M}_N$ and the closed-form congruence map $A_p$ (Eq.~\eqref{eq:Ap}), which transports only $P_N$ and $P_S$. (b) Schematic tangent-space view: the N and S prototypes move to the patient's normal-derived position, whereas the V and F prototypes stay at the source; the fixed $P_F$ acts as a geometric abstention buffer that absorbs borderline normal beats as F instead of V. Applying the same map to $P_V$ (\arm{GA\_All}) would move it away from the patient's actual V beats (Section~\ref{sec:geom}).}
\label{fig:pipeline}
\end{figure*}

To address these limitations, the main contributions of this study are as follows:
\begin{enumerate}
\item \textit{Normal-Anchored Geometric Transport (NAGT / \arm{GA\_NS}):} we propose an adaptation technique that estimates a patient-specific SPD congruence map from label-free normal anchors in the first 5-min prefix, applies it only to the supraventricular classes that share the normal conduction pathway (N, S), and fixes the ventricular (V) and fusion (F) prototypes at their source positions. The adaptation is computed in closed form, requires no back-propagation and changes no network parameter (Fig.~\ref{fig:pipeline}a,b).
\item \textit{Rate-invariant relative-RR branch and score offsets:} by combining dimensionless local relative-interval features (relA, adopted after INCART/SVDB were opened) with fixed score offsets determined only within MIT-BIH ($\delta_S=1$, $\delta_F=0.5$), the method resolves the sinus-tachycardia S false positives induced by absolute RR features and the heart-rate drift of long Holter recordings.
\item \textit{Error-flow analysis and identification of the geometric abstention buffer:} we quantify the mechanisms by which moving the V prototype causes V$\to$S and V$\to$N leakage and moving the F prototype causes the N$\to$V balloon effect, and we show that the F prototype fixed at the source acts as a geometric abstention buffer that absorbs ambiguous borderline beats and protects V precision.
\item \textit{Large-scale one-shot confirmatory evaluation on held-out chest-patch ECG and transparent development progression:} the locked final model was evaluated once on the held-out Icentia11k cohort (single-channel chest patch, 493 patients) under a pre-specified gatekeeping and Holm-adjusted testing hierarchy, confirming statistical superiority over four literature SFDA baselines and three geometric controls, with all intermediate development and negative ablation stages reported transparently (\hyperref[tab:history]{Supplementary Table~1}).
\item \textit{Post hoc mechanistic and robustness analyses of the frozen model:} evaluating the locked model outputs, we show that on the confirmatory cohort the adaptation gain is almost entirely patient-specific rather than a shared device offset, that it remains robust to unclassified prefix beats, fiducial jitter and annotation-free R-peak detection on Holter recordings, and that it translates into a marked reduction in false alarms per hour (Section~\ref{sec:posthoc_results}; Supplementary Tables~3--6).
\end{enumerate}

\section{Results}
\label{sec:results}

\NAGTtableMain

\subsection{One-Shot Confirmatory Evaluation on the Held-Out Icentia11k Chest-Patch Cohort}
\label{sec:icentia}
\begin{figure*}[!t]
\centering
\includegraphics[width=\textwidth]{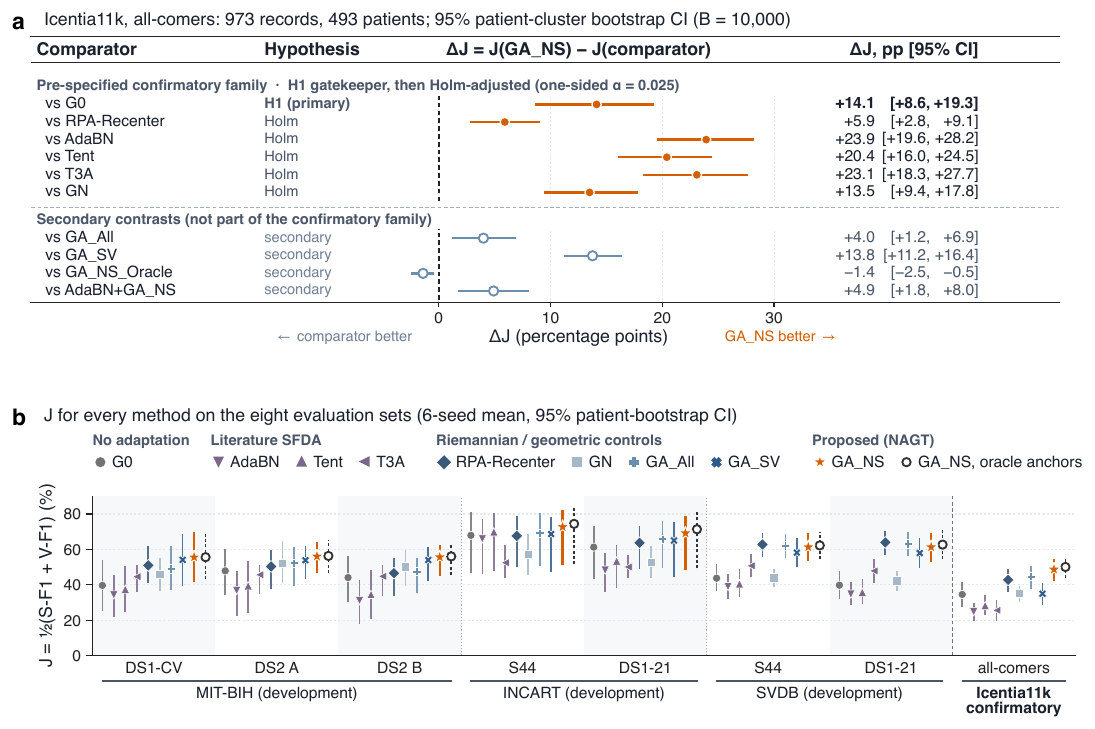}
\caption{Main benchmark. (a) Pre-specified one-shot confirmatory evaluation on the held-out Icentia11k cohort (all-comers; 973 records, 493 patients): $\Delta J=J(\text{\arm{GA\_NS}})-J(\text{comparator})$ with 95\% patient-cluster bootstrap CI ($B=10{,}000$). Filled circles (upper block): the gatekeeping hypothesis H1 (vs \arm{G0}) and the five Holm-adjusted contrasts (one-sided $\alpha=0.025$), all of which were confirmed. Hollow circles (below the dashed rule): secondary contrasts. The right-hand column lists $\Delta J$ [95\% CI] in percentage points. (b) $J$ (\%) with 95\% CI for all methods on the eight main evaluation sets (6-seed mean; final configuration for every method). MIT-BIH, INCART and SVDB are development sets; only Icentia11k (right of the dashed line) is confirmatory. DS2 A, DS2 B: MIT-BIH DS2 with Track A (14-patient) and Track B (DS1-21) sources; S44, DS1-21: MIT-BIH-44 and DS1-21 sources. \arm{AdaBN+GA\_NS} applies \arm{GA\_NS} prototype transport on top of AdaBN-updated features.}
\label{fig:benchmark}
\end{figure*}

In the Icentia11k all-comers cohort (493 patients, 973 records, 3,615,388 evaluated beats in total: 3,565,024 N, 23,076 S and 27,288 V), NAGT (\arm{GA\_NS}) achieved an arrhythmia macro-F1 $J=\tfrac{1}{2}(\text{S-F1}+\text{V-F1})=48.7\%$ [42.1, 54.7] (S-F1 $33.8\%$, S-Se $66.9\%$, S-PPV $22.6\%$, V-F1 $63.5\%$, V-Se $73.5\%$, N-Se $93.9\%$) (Table~\ref{tab:main}, \hyperref[tab:full]{Supplementary Table~2} and Fig.~\ref{fig:benchmark}a). The primary hypothesis H1 was confirmed ($\Delta J=+14.1$ [+8.6, +19.3]\pp\ against \arm{G0}, $34.5\%$ [27.4, 41.8]), and the five subsequent Holm multiple-comparison contrasts were all confirmed as well:
\begin{itemize}
\item vs \arm{RPA-Recenter} ($42.8\%$): $+5.9$ [+2.8, +9.1]\pp
\item vs \arm{AdaBN} ($24.7\%$): $+23.9$ [+19.6, +28.2]\pp
\item vs \arm{Tent} ($28.3\%$): $+20.4$ [+16.0, +24.5]\pp
\item vs \arm{T3A} ($25.6\%$): $+23.1$ [+18.3, +27.7]\pp
\item vs \arm{GN} ($35.2\%$): $+13.5$ [+9.4, +17.8]\pp
\end{itemize}
Against the auxiliary geometric controls \arm{GA\_All} ($44.7\%$, $\Delta J=+4.0$ [+1.2, +6.9]\pp) and \arm{GA\_SV} ($34.9\%$, $\Delta J=+13.8$ [+11.2, +16.4]\pp), \arm{GA\_NS} was also superior with confidence intervals above 0, and its difference from \arm{GA\_NS\_Oracle} ($50.1\%$), whose anchors were built from the true N beats, was only $-1.4$ [$-2.5$, $-0.5$]\pp. In the non-AF cohort (470 patients, 919 records), the $J$ of \arm{GA\_NS} rose to $59.8\%$ [53.9, 64.8] (S-F1 $49.7\%$, V-F1 $69.9\%$, N-Se $94.9\%$), a large gain of $+15.9$ [+9.6, +22.4]\pp\ over \arm{G0} and $+21.9$ [+17.6, +26.0]\pp\ over \arm{GN}. In the 73-patient held-out benchmark subset (Icentia test-75; 145 records), $J$ was $47.1\%$ [30.6, 63.8]; the superiority over \arm{AdaBN} ($+25.1$\pp), \arm{Tent} ($+19.5$\pp), \arm{T3A} ($+16.9$\pp) and \arm{GA\_SV} ($+10.8$\pp) was significant, but, given the small size of the subgroup, the contrasts with \arm{G0} ($+6.1$ [$-8.5$, +23.4]\pp) and \arm{GA\_All} ($-0.0$ [$-8.6$, +10.3]\pp) did not reach statistical significance.

\noindent\textbf{Clinical operating characteristics (post hoc).} Post hoc accounting of the frozen Icentia11k outputs (Section~\ref{sec:posthoc}; Supplementary Tables~3a--e) expresses these gains in triage terms. Over 1,052.5\,h of evaluated recording, \arm{GA\_NS} produced 13.22 ventricular false alarms (N$\to$V) per hour against 41.12 for \arm{G0} (per-record median 1.23/h [IQR 0.15, 4.93]) and 58.80 against 118.30 supraventricular-plus-ventricular false alarms per hour. Its one-vs-rest positive likelihood ratios were 45.5 for S and 181.5 for V (\arm{G0}: 31.6 and 65.3), and its three-class balanced accuracy over N, S and V was $78.10\%$ (\arm{G0} $69.86\%$). Record-level PVC burden followed the reference more closely (Pearson $r=0.700$ vs 0.447; mean absolute error 0.68 vs 1.38\pp). At the patient level, among the 451 patients for whom $J$ was defined under both models, \arm{GA\_NS} raised $J$ in 272 and left it unchanged in 92 (together 364, 80.7\%) and lowered it in 87 (median $\Delta J=+2.69$\pp).

\subsection{Cross-Database Benchmark and Behavior of BN- and Entropy-Based SFDA}
\label{sec:sfda}
On the development databases, \arm{GA\_NS} reached $55.6\%$ on MIT-BIH DS1-CV (non-AF $64.3\%$), $56.0\%$ / $55.6\%$ on DS2 Track A/B (non-AF $54.1\%$ / $58.3\%$), $72.7\%$ / $69.1\%$ on INCART (two sources; non-AF $73.8\%$ / $70.6\%$) and $61.3\%$ / $61.3\%$ on SVDB (two sources) (Table~\ref{tab:main}, \hyperref[tab:full]{Supplementary Table~2} and Fig.~\ref{fig:benchmark}b). Among the literature SFDA methods, the strongest comparator was \arm{RPA-Recenter}. \arm{GA\_NS} significantly outperformed \arm{RPA-Recenter} on MIT-BIH DS2 (Track A $+5.7$ [+1.7, +11.0]\pp, Track B $+9.1$ [+1.6, +15.5]\pp) and showed the same direction of superiority on INCART ($+5.1$ [$-2.7$, +11.1]\pp{} and $+5.4$ [$-2.1$, +9.8]\pp), whereas there was no significant difference on SVDB (MIT-BIH-44 source $-1.5$ [$-6.8$, +3.4]\pp, DS1-21 source $-2.7$ [$-8.2$, +2.5]\pp).

In contrast, \arm{AdaBN} and \arm{Tent} fell below the non-adapted \arm{G0} on MIT-BIH and SVDB. On Icentia11k in particular, the V sensitivity (V-Se) of the two methods rose to $92.0\%$ and $91.7\%$, respectively, whereas the V precision (V-PPV) dropped to $14.4\%$ and $15.6\%$ and the normal-beat specificity (N-Se) to $66.6\%$ and $68.7\%$, showing that many normal beats were pushed into the ventricular class. \arm{T3A} collapsed differently: as the S template adjusted from a few pseudo-labeled prefix beats became contaminated, its S-PPV on Icentia11k was only $5.6\%$. A plausible account of the collapse of the BN and entropy methods is the \emph{class-mixture mismatch} hypothesis: the BN statistics and entropy targets learned on class-balanced source batches (25\% each) are refitted to target prefixes in which 87.9--98.6\% of the beats are normal. A post hoc oracle intervention supports this account but also shows its limit (Supplementary Table~5c): re-estimating the BN statistics with the true prefix labels so that every class present carried equal weight (\emph{class-weighted AdaBN}; evaluation beats untouched) on the development records whose prefix contained more than one class (MIT-BIH DS2 A/B and INCART with both sources; 14--62 records per set) raised V-PPV over \arm{AdaBN} by 3.6 to 12.6\pp, yet $J$ remained below \arm{G0} in all four sets (e.g., INCART, DS1-21 source: $57.38\%$ vs $63.78\%$). A single per-channel BN renormalization therefore cannot undo a class-dependent shift even when the class mixture is corrected.

Output-only corrections that leave the network untouched did not close the gap on the confirmatory cohort either (Supplementary Tables~5a,b). EM re-estimation of each record's class prior from its prefix (Prior-EM\cite{saerens2002adjusting}; uniform source prior, as implied by balanced training) lowered $J$ on Icentia11k to $29.59\%$ and drove the estimated S prior below $10^{-6}$ in 87.3\% of record\,$\times$\,seed units, including 712 of the 1,068 units whose prefix truly contained S beats. Laplacian-regularized posteriors (LAME\cite{boudiaf2022lame}) reached $33.56\%$ with 64-beat online batches and $37.25\%$ even when the whole evaluation segment was used transductively, against $48.66\%$ for \arm{GA\_NS}. The output-only methods were not uniformly inferior, however: on INCART, where V beats make up 11.5\% of the evaluated beats, Prior-EM ($75.97\%$ / $75.58\%$) and LAME ($73.95\%$ / $69.71\%$) exceeded \arm{GA\_NS} ($72.70\%$ / $69.14\%$) with the MIT-BIH-44 / DS1-21 sources, whereas on MIT-BIH DS1-CV, DS2 and SVDB every variant stayed below \arm{GA\_NS}.

\NAGTtableAblation

\subsection{Stepwise Ablation and Error-Flow Analysis}
\label{sec:ablation}
\begin{figure*}[!t]
\centering
\includegraphics[width=\textwidth]{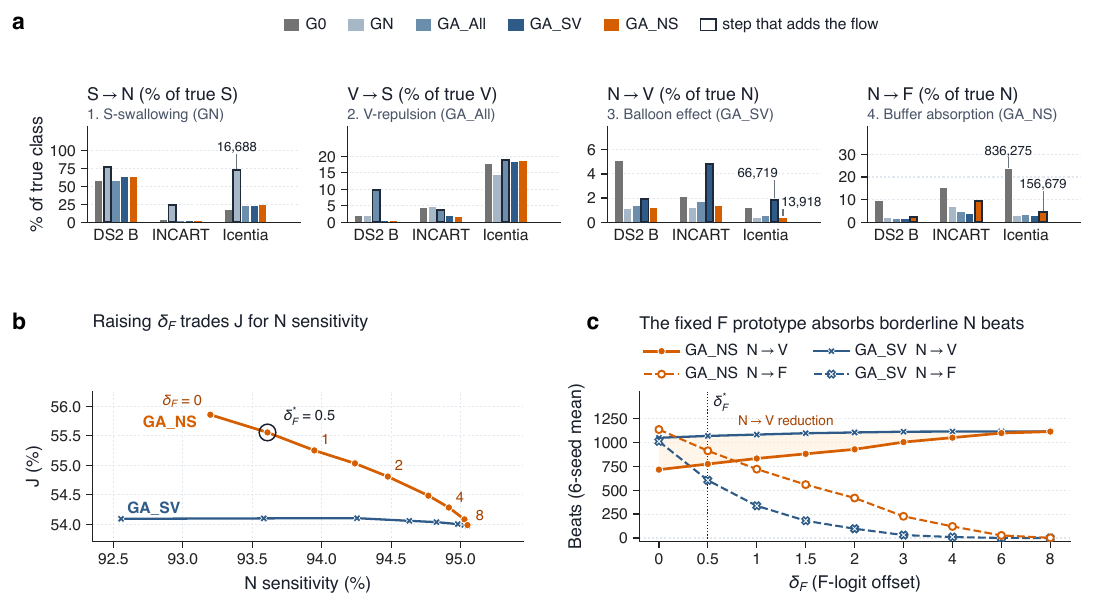}
\caption{Error flows and the geometric abstention buffer. (a) Error flows removed or created by each design step, as a percentage of the true class, on MIT-BIH DS2 (Track B source), INCART (MIT-BIH-44 source) and Icentia11k; final RR configuration (relA, $\delta_S=1$, $\delta_F=0.5$) for every arm, 6-seed means (counts in Table~\ref{tab:ablation}; Section~\ref{sec:ablation} additionally quotes the original four-dimensional RR-head counts of Table~\ref{tab:ablation}b,c). Outlined bars mark the step at which each flow appears: \arm{GN} swallows S beats (S$\to$N), \arm{GA\_All} repels V (V$\to$S), \arm{GA\_SV} inflates N$\to$V, and \arm{GA\_NS} converts borderline N$\to$V into N$\to$F. Selected Icentia11k counts are given with leader lines. (b) DS1-CV grid of the F-logit offset $\delta_F$ (selected value circled): $J$ versus N sensitivity for \arm{GA\_NS} and \arm{GA\_SV}. (c) N$\to$V and N$\to$F counts on DS1-CV as $\delta_F$ increases (shaded area: N$\to$V gap between \arm{GA\_SV} and \arm{GA\_NS}); when F is effectively masked (large $\delta_F$), \arm{GA\_SV} and \arm{GA\_NS} coincide.}
\label{fig:flow}
\end{figure*}

Table~\ref{tab:ablation}a--d and Fig.~\ref{fig:flow}a show how each design element blocks a specific type of misclassification.
\begin{itemize}
\item Normal-only re-centering (\arm{GN}) swallows S beats (S-swallowing): on MIT-BIH DS2 B (original four-dimensional RR-head stage, Table~\ref{tab:ablation}b), the S$\to$N errors increased from 963 for \arm{G0} to 1,176 for \arm{GN} and S-Se dropped from $22.8\%$ to $16.3\%$; on Icentia11k (final configuration, Table~\ref{tab:ablation}d), S$\to$N surged from 3,837 to 16,688 and S-Se plunged from $61.0\%$ to $18.4\%$.
\item Transporting all four prototypes (\arm{GA\_All}) repels the V prototype: on DS2 B (Table~\ref{tab:ablation}b), V$\to$S leakage jumped to 360 for \arm{GA\_All} against 33 for \arm{GA\_NS}.
\item Even with V fixed, transporting F as well (\arm{GA\_SV}) produces the N$\to$V balloon effect: the N$\to$V false positives of \arm{GA\_SV} against \arm{GA\_NS} rose to 792 vs 306 on DS2 B and 5,863 vs 2,143 on INCART (both at the original four-dimensional RR-head stage, Table~\ref{tab:ablation}b,c), and 66,719 vs 13,918 on Icentia11k (final configuration, Table~\ref{tab:ablation}d; V-PPV $24.2\%$ vs $56.9\%$, V-F1 $36.00\%$ vs $63.55\%$).
\item By fixing $P_F$ at the source, \arm{GA\_NS} makes borderline normal beats fall into F instead of V: on Icentia11k, N$\to$F increased to 156,679 (\arm{GA\_SV}: 104,992), while the total N$\to$V\,+\,N$\to$F is almost identical for the two models (170,597 vs 171,711). If the F class is masked so that it cannot be output, \arm{GA\_SV} and \arm{GA\_NS} become mathematically identical models and $J$ drops sharply (Icentia11k $33.96\%$; MIT-BIH DS1-CV $53.98\%$ vs $55.56\%$). Post hoc accounting shows that the buffer rarely captures true arrhythmias: on Icentia11k only 0.54\% of the true S and 3.52\% of the true V beats were output as F (Supplementary Table~3a). Because merging the F output into the non-alarm class leaves every S and V count unchanged, reading F as a non-alarm abstention keeps $J=48.66\%$ and the S and V sensitivities and precisions exactly, while the effective non-alarm specificity ((N$\to$N\,+\,N$\to$F)/N) is $98.26\%$ against the standard four-class N-Se of $93.87\%$ (\arm{G0}: $96.51\%$ vs $73.05\%$; Supplementary Table~3b).
\item The rate-invariant relative-RR head (relA) removed most of the N$\to$S false positives caused by sinus tachycardia (INCART total N$\to$S: 18,146 $\to$ 1,269 [final model with $\delta_F=0.5$: 1,281]).
\end{itemize}

\NAGTtableGeom

\subsection{Geometric Invariance across Four Databases}
\label{sec:geom}
\begin{figure*}[!t]
\centering
\includegraphics[width=\textwidth]{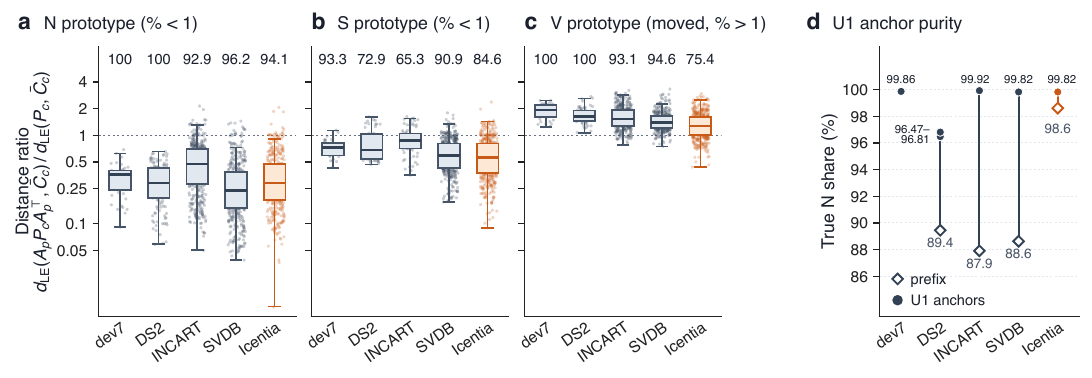}
\caption{Geometric invariance of the normal-derived map across databases. (a)--(c) Distance ratio $d_{\mathrm{LE}}(A_pP_cA_p^{\top},\bar{C}_c)/d_{\mathrm{LE}}(P_c,\bar{C}_c)$ per evaluation unit (record\,$\times$\,seed with $\geq 10$ evaluation beats of the class), where $\bar{C}_c$ is the record's evaluation class-$c$ mean; a ratio $<1$ means that the map moves the prototype towards the patient's own class-$c$ beats. Numbers above the boxes are the percentages of units in the expected direction (N, S: $<1$; V: $>1$, i.e., the V prototype would be repelled if it were transported). Boxes: median and IQR; whiskers: 1.5\,IQR; points: individual units (at most 400 shown per box). dev7: MIT-BIH development patients (seeds 0--5); DS2: Track B source; INCART and SVDB: MIT-BIH-44 source; Icentia: Icentia11k (ratios equal to 1 are records with identity maps). (d) U1 anchor purity (share of true N among the selected anchors; for DS2 the two track values are drawn as circles joined by a bar) and the N share of the label-free prefix (vertical stems connect the prefix N share with the anchor purity) (labels used for description only; dev7 purity from seeds 0--2).}
\label{fig:geom}
\end{figure*}

As Table~\ref{tab:geom} and Fig.~\ref{fig:geom} show, the map $A_p$ estimated from the normal anchors moved the N prototype closer to the mean of the patient's actual N beats in 92.9--100\% of the evaluation units (record\,$\times$\,seed) of every database (Icentia11k $94.1\%$, median distance ratio $0.290$), and the S prototype also came closer to the mean of the actual S beats in 65.3--100\% of the units (Icentia11k $84.6\%$, median distance ratio $0.560$). Conversely, the same map pushed the V prototype away from the patient's actual V beats in 92.8--100\% of the units of MIT-BIH, INCART and SVDB (median ratio 1.394--2.14), and in $75.4\%$ of the units on the Icentia11k chest patch (median ratio $1.281$). The true normal-beat purity of the label-free U1 anchors reached 96.5--99.9\% in every database (Icentia11k $99.82\%$).

\NAGTtableRuntime

\subsection{Heart-Rate Drift and Rhythm Stratification}
\label{sec:drift}
With the original absolute RR features, the S false-positive rate of the INCART records closely followed the heart rate (Spearman $\rho=-0.61$ with the median normal RR): the six fast-rate records with a median normal RR below 0.60\,s contained only 12\% of all normal beats but produced 32--39\% of all N$\to$S errors. relA with the $\delta_S=1$ correction reduced the N$\to$S false positives of these six INCART records from 7,110 to 90 and those of MIT-BIH record 215 from 1,020 to 18 (evaluated at $\delta_F=0$; Fig.~\ref{fig:rate}a). The prefix-normalized variant relB, on the other hand, failed on records whose baseline heart rate rose after the prefix (INCART I18: 279 vs 8 for relA; I22: 686 vs 48). Indeed, the within-record heart-rate drift (90th percentile of $|\log(m_i/m_{\mathrm{prefix}})|$) was more than twice as large in INCART ($0.334$) as in MIT-BIH DS2 ($0.151$). Moreover, in the non-AF cohort, which excludes the AF recordings, the gain of \arm{GA\_NS} over \arm{G0} on Icentia11k widened further to $+15.9$ [+9.6, +22.4]\pp.

\subsection{Adaptation Cost}
\label{sec:cost}
\begin{figure*}[!t]
\centering
\includegraphics[width=\textwidth]{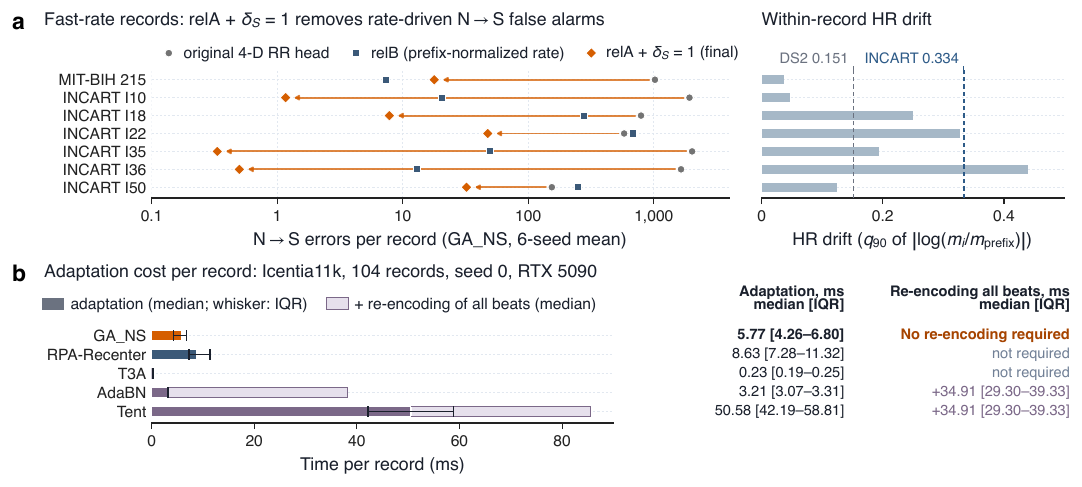}
\caption{Rate invariance and adaptation cost. (a) N$\to$S errors of \arm{GA\_NS} on fast-rate records (median normal RR $<0.60$\,s; MIT-BIH 215 and six INCART records, MIT-BIH-44 source) with the original four-dimensional RR head, the prefix-normalized relB head and the final relA head with $\delta_S=1$ ($\delta_F=0$; 6-seed means; values below 0.1 are plotted at 0.1). Right: within-record heart-rate drift (90th percentile of $|\log(m_i/m_{\mathrm{prefix}})|$ over the evaluation N beats); dashed lines: database-level values for MIT-BIH DS2 (0.151) and INCART (0.334). (b) Adaptation time per record on Icentia11k (104 records, seed 0; median and IQR; NVIDIA RTX 5090 for network operations, NumPy with 8 threads for SPD operations). \arm{AdaBN} and \arm{Tent} additionally require re-encoding all beats of the record (lighter segment; median and IQR of each component are listed on the right).}
\label{fig:rate}
\end{figure*}

As Table~\ref{tab:runtime} and Fig.~\ref{fig:rate}b show, \arm{GA\_NS} reuses the frozen encoder features as they are and completes the adaptation in a median of 5.77\,ms [IQR 4.26--6.80\,ms] per record. \arm{Tent}, in contrast, requires 50.58\,ms of back-propagation adaptation plus an additional 34.91\,ms to re-encode the whole record with the updated BN layers.

\subsection{Decomposition of Patient Specificity and Front-End Robustness (Post Hoc)}
\label{sec:posthoc_results}
Two sets of post hoc analyses of the frozen model (Section~\ref{sec:posthoc}) evaluate whether the confirmed gain is patient-specific and whether it remains robust to imperfect front-end inputs.

\noindent\textbf{Dataset-level versus patient-specific shift.} Replacing each record's own map $A_p$ by a leave-one-patient-out dataset-global map ($A_{\mathrm{device}}$: built from the AIRM mean of the U1 anchor means of all other patients of the same cohort, so that it captures the offset shared by the device and lead) recovered only $+0.50$ [$-1.59$, $+2.60$]\pp\ (3.5\%) of the $+14.12$\pp\ Icentia11k gain; the remaining $+13.62$ [$+8.21$, $+18.59$]\pp\ (96.5\%) required the patient's own anchors (Supplementary Table~4a). On MIT-BIH DS2 with the DS1-21 source, the patient-specific component was $+11.47$ of $+11.54$\pp\ (99.4\%). The split was not the same in every development database: the dataset-level component accounted for 25.8\% of the gain on DS2 with the 14-patient source, 35.7\% on SVDB and 72.3\% on INCART (MIT-BIH-44 sources). Two null controls exclude a geometric coincidence on Icentia11k: assigning each record the map of a randomly drawn other patient lowered $J$ by 4.74\pp\ relative to \arm{G0} (N distance-ratio median 1.114, only 32.7\% of units below 1), and a random congruence of the same tangent-space magnitude lowered it by 13.13\pp\ (N ratio below 1 in 0.17\% of units; Supplementary Table~4b).

\noindent\textbf{Unclassified beats in the prefix.} The frozen protocol excludes Icentia11k beats labeled Q from the prefix candidates. When all Q beats of the first 5\,min were instead encoded and passed unchanged into the U1 selection (74,845 Q beats per seed; evaluation segment unchanged), the RR filter removed 47.5\% of them and the 80\% LE core a further 17.1\%, so the two label-free stages discarded 64.6\%; the remaining Q beats made up 13.8\% of the anchors (true-N purity $86.0\%$ instead of $99.82\%$). $J$ was nevertheless unchanged ($48.66\%\to48.67\%$; non-AF $+0.21$\pp; \arm{RPA-Recenter}, which uses every prefix beat, $+0.94$\pp), and 31 of the 35 records whose frozen prefix was empty obtained a prefix and could be adapted (pooled $J$ of these 35 records $51.83\%\to51.08\%$; Supplementary Tables~6a,b).

\noindent\textbf{Fiducial jitter and annotation-free detection.} Displacing every realigned fiducial by uniform jitter of $\pm10$ and $\pm25$\,ms preserved $99.34\%$ and $97.31\%$ of the Icentia11k $J$ of \arm{GA\_NS}, more than for \arm{G0} ($97.68\%$, $93.64\%$) and \arm{RPA-Recenter} ($98.66\%$, $96.23\%$); at $\pm50$\,ms retention fell to $85.56\%$ (\arm{G0} $75.01\%$, \arm{RPA-Recenter} $79.64\%$; Supplementary Table~6c). On the Holter databases, running the whole pipeline from R peaks found by the open-source \texttt{xqrs} detector, without any annotation position (beat-detection sensitivity 98.20--99.68\%), preserved the adaptation gain: on MIT-BIH DS2 (DS1-21 source) and INCART (MIT-BIH-44 source), \arm{GA\_NS} reached $J=53.85\%$ and $67.02\%$ ($-1.76$ and $-5.68$\pp\ relative to annotation fiducials) and exceeded \arm{G0} by $+11.18$ and $+4.11$\pp\ (annotation-driven: $+11.54$ and $+4.75$\pp). Across all six Holter evaluation sets, the gain over \arm{G0} of the detector-driven pipeline ($+4.11$ to $+19.32$\pp) remained close to that of the annotation-driven pipeline ($+4.75$ to $+21.50$\pp); the results for every database are listed in Supplementary Tables~6d,e.

\section{Discussion}
\label{sec:discussion}
\noindent\textbf{Electrophysiological interpretation and its scope.} The class-wise transport directions are consistent with the conduction-pathway argument: the map estimated from normal beats transfers to supraventricular beats but not to ventricular ectopic beats. This evidence is at the level of macroscopic correlations, but a post hoc split of the supraventricular class by PhysioNet beat symbol sharpens it (Supplementary Table~4c): on MIT-BIH DS2 the normal-derived map moved the S prototype towards normally conducted atrial, nodal and supraventricular premature beats (A, J, S; median distance ratio 0.732 and 0.646 for the 14-patient and DS1-21 sources, 85.7\% and 88.1\% of units below 1) but away from aberrated atrial premature beats (a; median 1.204 and 1.415, 83.3\% and 100\% of units above 1), which, like V beats, depolarize the ventricles over an abnormal route. Because the aberrated subgroup is small (two records, 12 record\,$\times$\,seed units) and the S transport direction was more variable on INCART (65.3--72.2\%), finer stratification by conduction pathway and ectopic focus in multi-lead recordings remains a subject for future study.

\noindent\textbf{Why the anchor-selection head is kept four-dimensional.} Unifying the architecture by re-selecting the U1 anchors with the rate-invariant classification head (relA) was rejected under our pre-specified MIT-BIH decision rule (\hyperref[tab:history]{Supplementary Table~1}): on DS1-CV the anchor purity dropped by 0.55\pp, exceeding the 0.50\pp\ tolerance, and with the MIT-BIH-44 source the number of prefix anchors passed by the relA head fell by roughly half. This reflects the different requirements of the two stages: anchor selection within a stationary 5-min prefix benefits from conservative normal-beat filtering, whereas classification over subsequent hours requires invariance to heart-rate drift. Unifying both roles in a single head is left for future work.

\noindent\textbf{The F prototype as a geometric abstention buffer and the reporting of N-Se.} The untransported F prototype absorbs borderline beats that would otherwise have become V false positives, which considerably raises V-PPV and $J$. Because $J$ does not count F errors, however, the absorbed beats appear as N$\to$F (4.4\% of all N beats for \arm{GA\_NS} on Icentia11k) and lower the overall normal-beat specificity (N-Se). Readers comparing methods in terms of total normal specificity should therefore consult the N-Se that this paper reports in every table; read as a non-alarm abstention, F keeps 98.26\% of normal beats out of the alarm classes (Section~\ref{sec:ablation}). On MIT-BIH, reducing the buffer size (increasing $\delta_F$) showed a direct trade-off between $J$ and N-Se (Fig.~\ref{fig:flow}b,c), and masking F completely removed the advantage of \arm{GA\_NS}.

\noindent\textbf{Weaker V repulsion on the chest-patch lead.} On the Icentia11k chest patch, the share of units in which the normal-derived map pushed the V prototype away was 75.4\% (median ratio 1.281), milder than on the other three databases (MIT-BIH, INCART, SVDB; 92.8--100\%). A post hoc analysis points to the reduced N--V morphological contrast of the modified lead I patch lead (Supplementary Table~4d): across the 1,092 Icentia11k record\,$\times$\,seed units with at least 10 evaluated V beats, the V distance ratio increased with the record's baseline N--V distance $d_{\mathrm{LE}}(\bar{C}_N,\bar{C}_V)$ (Spearman $\rho=0.229$, $p=1.9\times10^{-14}$), and the share of units in which the V prototype would be repelled rose from 61.9\% in the lowest to 87.2\% in the highest quartile of that distance. Even so, \arm{GA\_NS} achieved the highest V-F1 among all non-oracle methods on Icentia11k ($63.55\%$, while simultaneously preserving S-F1 at $33.77\%$ vs $6.83\%$ for \arm{GN}; Table~\ref{tab:ablation}d and \hyperref[tab:full]{Supplementary Table~2}).

\noindent\textbf{Rhythm boundaries and unclassified (Q) beats.} Supraventricular premature beat (S) detection is not defined during atrial fibrillation (AF), so the non-AF cohort is reported alongside every analysis. In addition, 24.5\% of all beat events in Icentia11k are annotated as unclassified (Q) and are excluded from both the prefix candidates and the evaluation targets under the primary protocol; however, our post hoc sensitivity analysis admitting all prefix Q beats into anchor selection left $J$ unchanged ($48.66\%\to48.67\%$; Section~\ref{sec:posthoc_results}), confirming that prefix adaptation does not depend on prior Q-beat filtering. Furthermore, at an S prevalence of only 0.64\%, the absolute S-PPV (22.6\%) remains modest, and the relA head with $\delta_S=1$ entails a sensitivity--precision trade-off (evaluated at $\delta_F=0$): on SVDB (MIT-BIH-44 source), S-PPV rose from $31.2\%$ to $63.6\%$ while S-Se fell from $71.9\%$ to $40.9\%$, and on DS1-CV S-PPV rose from $22.2\%$ to $27.1\%$ while S-Se fell from $66.6\%$ to $33.3\%$.

\noindent\textbf{Transparency of the research process.} The final model was reached through a sequence of development stages on the development cohorts, followed by a single confirmatory evaluation (\hyperref[tab:history]{Supplementary Table~1}). The architecture was first tuned on the seven MIT-BIH development patients (dev7), and the V-fixed variant \arm{GA\_SV} was then confirmed in a locked evaluation on new encoder seeds of the same patients (positive). The subsequent evaluation on MIT-BIH DS2 was negative against \arm{GN}, after which \arm{GA\_NS} and its fixed-F buffer were derived post hoc on DS1 (DS1-CV). The evaluation of the original four-dimensional-head model on INCART was negative, and the rate-invariant relA head was adopted after the INCART and SVDB results had been seen. Finally, the locked final model, with all model weights, hyperparameters and decision rules fixed, was evaluated once on the held-out Icentia11k cohort (positive). The MIT-BIH DS2, INCART and SVDB results are therefore development results that reflect the method-development process, and only Icentia11k is a strict confirmatory test.

\noindent\textbf{Conclusion.} The label-free first 5-min prefix alone suffices to estimate a patient-specific SPD congruence map that adapts the normal and supraventricular prototypes to the patient while keeping the ventricular and fusion prototypes at their source positions. In 493 single-channel chest-patch patients, this closed-form, gradient-free adaptation significantly outperformed the non-adapted model, normal-only and non-anchored Riemannian re-centering, and test-time adaptation methods based on batch normalization, entropy and templates.
\label{sec:conclusion}

\section{Methods}
\label{sec:method}

\subsection{Problem Formulation}
\label{sec:problem}
The source classification model is trained on a labeled source population (the 44 non-paced records of the MIT-BIH Arrhythmia Database\cite{moody2001impact,goldberger2000physionet}; 43 unique subjects after records 201 and 202, which come from the same subject, are merged), after which its weights are completely frozen, and no source data are accessible at target-adaptation time. For each target recording, the first 5\,min of unlabeled prefix are observed; the algorithm adapts using this segment only and must then classify every beat of the subsequent evaluation segment into one of the four AAMI standard classes $c\in\{\mathrm{N},\mathrm{S},\mathrm{V},\mathrm{F}\}$\cite{aami2012ec57}. Beat fiducial positions are taken from the annotation files and realigned with a fixed signal rule (Section~\ref{sec:protocol}); the true class labels of the target beats are used nowhere in adaptation or decision making.

\subsection{Dual-Branch Model and SPD Covariance Representation}
\label{sec:model}
Each beat is encoded by a dual structure: a convolution--covariance branch that processes the waveform morphology and a relative-RR branch that processes the beat intervals (Fig.~\ref{fig:pipeline}a).
In the morphology branch, a one-dimensional convolutional encoder (three Conv--BatchNorm--GELU blocks with 16, 32 and 32 channels followed by a $1\times 1$ convolution to $d=8$ channels; 6,248 parameters in total) maps a 288-sample waveform crop (from 0.3\,s before to 0.5\,s after the fiducial point, 360\,Hz sampling, causal 0.5--40\,Hz band-pass filter) to a feature map $H\in\mathbb{R}^{8\times 72}$ ($L=72$). The sample covariance is computed from $\tilde{H}$, in which the temporal mean of each channel has been removed, and trace normalization, shrinkage regularization ($\epsilon=0.001$) and eigenvalue floor clipping are applied to produce the $8\times 8$ symmetric positive-definite (SPD) morphology descriptor $C(x)\in\mathcal{S}_{++}^{8}$:
\begin{equation}
\begin{gathered}
\Sigma=\frac{1}{L-1}\tilde{H}\tilde{H}^{\top},\\
C(x)=(1-\epsilon)\,\frac{d\,\Sigma}{\mathrm{tr}(\Sigma)}+\epsilon I_d .
\end{gathered}
\label{eq:cov}
\end{equation}
The rhythm branch is a multilayer perceptron (MLP) that takes local relative-RR features as input (Section~\ref{sec:rr}). For rare-class learning, the encoder is trained under class-balanced sampling (1:1:1:1) together with a covariance proxy head and the RR head (AdamW, learning rate $10^{-3}$, 30 epochs, six independent seeds).

\subsection{Patient-Balanced Source Prototypes}
\label{sec:proto}
To prevent the waveforms of a few patients with frequent arrhythmias from dominating the source reference, the source prototype $P_c\in\mathcal{S}_{++}^{8}$ of each class $c$ is defined as a patient-balanced Fr\'echet mean under the affine-invariant Riemannian metric (AIRM)\cite{pennec2006riemannian}. That is, the AIRM mean $\mu_{k,c}$ of the class-$c$ beats within each source patient $k$ is computed first, and a second-level AIRM mean is then computed over the set $\mathcal{K}_c$ of patients that have the class:
\begin{equation}
P_c=\arg\min_{P\in\mathcal{S}_{++}^{8}}\ \sum_{k\in\mathcal{K}_c} d_{\mathrm{AIRM}}^{2}\!\left(P,\mu_{k,c}\right).
\label{eq:proto}
\end{equation}
During development, this beat-to-patient re-weighting improved the adaptation performance $J$ by +2.4 to +10.5\pp\ in all ten encoder snapshots (Table~\ref{tab:ablation}a, A3).

\subsection{Label-Free Normal Anchors (U1) and Closed-Form Transport Map}
\label{sec:anchor}
Within the first 5-min prefix of the target, beat $i$ becomes a primary anchor candidate when its current RR interval is not premature ($\Delta_i/m_i\geq 0.9$, where $m_i$ is the median of the preceding 10 intervals) and the original four-dimensional RR head assigns it a normal posterior probability $p(\mathrm{N})\geq 0.5$. Then, from the covariance cloud of the candidates, the 80\% of the candidates closest to the patient's own Log-Euclidean (LE) mean\cite{arsigny2007geometric} are retained as the final normal anchors (if fewer than 20 beats pass the RR stage, all prefix candidate beats enter the 80\%-core stage, and if there is no valid candidate beat at all, the identity map $A_p=I_d$ is applied). The AIRM mean $\hat{M}_N$ of the extracted normal anchors defines, in closed form and without back-propagation, the symmetric SPD congruence map $A_p$ that maps the source normal prototype $P_N$ to the target center $\hat{M}_N$ ($A_pP_NA_p^{\top}=\hat{M}_N$):
\begin{equation}
A_p=P_N^{-1/2}\left(P_N^{1/2}\hat{M}_NP_N^{1/2}\right)^{1/2}P_N^{-1/2}.
\label{eq:Ap}
\end{equation}
Eq.~\eqref{eq:Ap} has two equivalent readings. Algebraically, $A_p=P_N^{-1}\#\hat{M}_N$, where $X\#Y=X^{1/2}(X^{-1/2}YX^{-1/2})^{1/2}X^{1/2}$ is the matrix geometric mean, i.e.\ the midpoint of the AIRM geodesic between $P_N^{-1}$ and $\hat{M}_N$\cite{bhatia2009positive,moakher2005geometric}. It is also the linear optimal-transport map between the zero-mean Gaussian distributions $\mathcal{N}(0,P_N)$ and $\mathcal{N}(0,\hat{M}_N)$ under the 2-Wasserstein distance, whose closed form is the same symmetric matrix and whose cost is the Bures--Wasserstein distance between $P_N$ and $\hat{M}_N$\cite{dowson1982frechet,olkin1982distance,bhatia2019bures}. NAGT is therefore a class-selective optimal transport of covariance descriptors, related to optimal-transport domain adaptation\cite{courty2017optimal}.

\subsection{Class-Selective Transport (\arm{GA\_NS}) and the Geometric Abstention Buffer}
\label{sec:transport}
The proposed \arm{GA\_NS} applies the estimated congruence map $A_p$ only to the N and S prototypes, which share the normal conduction pathway, and keeps the V and F prototypes fixed at their source positions ($\alpha=1$):
\begin{equation}
\begin{gathered}
\tilde{P}_N=A_pP_NA_p^{\top},\quad \tilde{P}_S=A_pP_SA_p^{\top},\\
\tilde{P}_V=P_V,\quad \tilde{P}_F=P_F .
\end{gathered}
\label{eq:transport}
\end{equation}
The electrophysiological rationale is as follows: normal sinus beats (N) and supraventricular premature beats (S) both reach the ventricles through the atrioventricular node and the His--Purkinje conduction system, so their QRS morphologies share the patient-specific projection and conduction factors captured by the normal anchors. In contrast, premature ventricular contractions (V) are generated at an ectopic ventricular focus, and fusion beats (F) mix two conduction wavefronts, so there is no reason for a map derived from normal beats to fit them (see Section~\ref{sec:geom}, Section~\ref{sec:discussion} and Supplementary Table~4c for empirical validation). Furthermore, the F prototype ($P_F$), which is not transported and remains at its source position, acts as a \emph{geometric abstention buffer} in the joint decision space: ambiguous normal beats lying in the boundary region between the transported normal cloud ($\tilde{P}_N$) and the fixed source ventricular prototype ($P_V$) are absorbed as F instead of V, which blocks the critical N$\to$V false-positive balloon effect (Section~\ref{sec:ablation}, Fig.~\ref{fig:pipeline}b).
The two distances serve different purposes. Prototype transport, computed once per record, uses the AIRM because congruence is an AIRM isometry ($d_{\mathrm{AIRM}}(AXA^{\top},AYA^{\top})=d_{\mathrm{AIRM}}(X,Y)$ for any invertible $A$)\cite{pennec2006riemannian}, so the transported prototypes keep their mutual geometry exactly; the millions of beat-level scores of Eq.~\eqref{eq:score} use the Log-Euclidean distance, the Euclidean distance between matrix logarithms (the tangent space at the identity), which coincides with the AIRM distance for commuting matrices and avoids a matrix square root and inverse per beat\cite{arsigny2007geometric}.

\subsection{Rate-Invariant Relative RR and Score Fusion}
\label{sec:rr}
The RR head of the classification stage uses relA, a four-dimensional feature vector composed only of dimensionless local relative ratios:
\begin{equation}
\begin{aligned}
r_i^{\mathrm{relA}}=\Big[&\log\frac{\Delta_i}{m_i},\ \log\frac{\Delta_{i+1}}{m_i},\\
&\log\frac{\Delta_i}{\Delta_{i-1}},\ \log\frac{\Delta_{i+1}}{\Delta_i}\Big].
\end{aligned}
\label{eq:relA}
\end{equation}
These features do not change when all beat intervals are scaled by an arbitrary constant, so a fast sinus rhythm (tachycardia) is not misread as prematurity. A prefix-normalized alternative (relB), which adds a local-rate term relative to the prefix median RR ($\log(m_i/m_{\mathrm{prefix}})$) and RR-jitter features, was rejected because its false positives surged when the baseline heart rate drifted after the prefix (Section~\ref{sec:drift}). The final fused score $s_c(x)$ of class $c$ for an evaluation beat $x$ and the predicted class $\hat{y}$ are defined as
\begin{equation}
\begin{aligned}
s_c(x)&=-\frac{d_{\mathrm{LE}}^{2}\!\left(C(x),\tilde{P}_c\right)}{\tau\eta}+\ell_c^{\mathrm{RR}}(x)\\
&\quad-\delta_S\,\mathbb{I}[c=\mathrm{S}]-\delta_F\,\mathbb{I}[c=\mathrm{F}],\\
\hat{y}&=\arg\max_{c\in\{\mathrm{N},\mathrm{S},\mathrm{V},\mathrm{F}\}} s_c(x),
\end{aligned}
\label{eq:score}
\end{equation}
where $\tau=0.15$ is the temperature scale, $\eta$ is the source distance scale (median squared Log-Euclidean distance of the source training beats to their own-class prototype, fixed per source model), and $\delta_S=1$ and $\delta_F=0.5$ are fixed score offsets selected only within MIT-BIH. The theoretical correction that adds the full log class prior of the source training data to the logits (complete restoration of the prior removed by class-balanced training) performed worse for every head, so the empirical partial correction $\delta_S=1$ was adopted.

\subsection{Comparators}
\label{sec:comparators}
As geometric controls, we evaluated the non-adapted source model (\arm{G0}), \arm{GN}, which transports only N, \arm{GA\_All}, which transports all four prototypes, \arm{GA\_SV}, which fixes only V and transports N, S and F, and a diagnostic upper-bound model (\arm{GA\_NS\_Oracle}) that uses all true N beats of the prefix as anchors. Four representative literature SFDA baselines were implemented fairly on the same frozen encoder checkpoints and on the same relA RR head and offsets:
\begin{enumerate}
\item \arm{RPA-Recenter}: Riemannian re-centering that maps the AIRM mean of all prefix beats, without anchor selection or class distinction, onto the mean of all source beats and applies the map to all four prototypes\cite{zanini2018transfer,rodrigues2019rpa}.
\item \arm{AdaBN}: re-estimation of the running means and variances of the three BatchNorm layers of the encoder from the prefix waveform crops\cite{li2018adabn}.
\item \arm{Tent}: back-propagation minimization of the Shannon entropy of the fused score with respect to the BN affine parameters (160 parameters) (Adam, learning rate $10^{-3}$, 5 epochs, batch size 64)\cite{wang2021tent}.
\item \arm{T3A}: adjustment of the prototypes to the Log-Euclidean mean of the top $M=5$ prefix beats with the lowest-entropy pseudo-labels per class\cite{iwasawa2021t3a}.
\end{enumerate}
\arm{Tent} and \arm{T3A} are versions ported to our SPD prototype pipeline, and the hyperparameters of all SFDA baselines were selected only on MIT-BIH DS1-CV and then fixed.

\subsection{Databases}
\label{sec:setup}
\label{sec:data}
\begin{enumerate}
\item MIT-BIH Arrhythmia Database (MLII lead, 360\,Hz)\cite{moody2001impact,goldberger2000physionet}: following the DS1/DS2 inter-patient split\cite{dechazal2004automatic}, the 44 records that remain after excluding the paced records were used (records 201 and 202, from the same subject, were grouped, giving 43 unique subject clusters).
\item St.\ Petersburg INCART Database (standard lead II, 257\,Hz)\cite{tihonenko2008incart}: 75 30-min Holter recordings from 32 patients.
\item MIT-BIH Supraventricular Arrhythmia Database (SVDB; ECG1 lead, 128\,Hz)\cite{greenwald1990svdb}: 78 30-min recordings enriched in supraventricular arrhythmias.
\item Icentia11k Database (single-channel modified lead I chest patch, 250\,Hz)\cite{tan2022icentia,tan2021icentia}: from the 11,000 patients, 500 patients $\times$ 2 non-contiguous segments (about 70\,min per segment, 1,000 recordings in total) were selected via deterministic hash-based sampling of identifiers, independently of labels and signal quality; after applying the pre-specified data-exclusion criteria (Section~\ref{sec:prereg}), 973 recordings from 493 patients were analyzed.
\end{enumerate}
All signals are resampled to 360\,Hz with a polyphase filter.

\subsection{Evaluation Protocol and Statistical Analysis}
\label{sec:protocol}
Each recording is adapted independently from its own first 5-min prefix and evaluated on the remaining segment (about 25\,min for Holter and about 65\,min for the patch). Beat fiducial points are realigned in bulk with a label-agnostic rule fixed on MIT-BIH (search for the maximum of $|y|$ within $\pm 50$\,ms of the annotation position, then subtract 3 samples). The source models for external evaluation are MIT-BIH-44 (primary source; 43 clusters) and DS1-21 (auxiliary source), each trained with six seeds (for the MIT-BIH internal evaluations, DS2 A uses a 14-patient source, DS2 B the DS1-21 source, and DS1-CV fold sources that exclude the target patients). The primary metric $J=\tfrac{1}{2}(\text{S-F1}+\text{V-F1})$ is computed from the confusion matrix pooled over all records and averaged over the six seeds; because $J$ does not reflect F errors, the normal-class sensitivity (N-Se, i.e., the specificity of arrhythmia detection on normal beats) and the per-class sensitivity (Se) and positive predictive value (PPV) are reported in every table. The 95\% confidence intervals were computed by patient-cluster bootstrap ($B=10{,}000$) (records were treated as clusters for SVDB, which has no patient IDs). Cohorts are reported as all patients (all-comers) and non-atrial-fibrillation (non-AF: AF/AFL duration $<60$\,s in the evaluation segment); prematurity presupposes an organized atrial baseline rhythm, so S-beat detection is clinically defined only outside AF.

\subsection{Confirmatory Evaluation Protocol and Data-Handling Rules}
\label{sec:prereg}
Before model inference was run on Icentia11k, the statistical testing hierarchy and two data-handling rules were locked, and the model weights, hyperparameters and decision rules developed on the development cohorts were fixed. In the testing hierarchy, the gatekeeping hypothesis H1: $J(\text{\arm{GA\_NS}})-J(\text{\arm{G0}})>0$ (one-sided $\alpha=0.025$, lower bound of the 95\% CI $>0$) was tested first, and, if H1 passed, Holm step-down multiple-comparison correction at one-sided $\alpha=0.025$ was applied to the five comparators \arm{RPA-Recenter}, \arm{AdaBN}, \arm{Tent}, \arm{T3A} and \arm{GN}. The two data-handling rules, fixed before any model output was generated, were: (1) removal of 7 duplicate annotations at identical sample indices in 5 records, which violated the monotonic beat ordering required for realignment; and (2) exclusion from the analysis of 27 records with no valid annotated beat other than unclassified Q beats, while the 35 records with an empty 5-min prefix defaulted to the non-adapted source model (\arm{G0}). The complete development history, including the negative MIT-BIH DS2 and INCART (original-head model) results, is reported transparently in the Supplementary Information (\hyperref[tab:history]{Supplementary Table~1}).

\subsection{Post Hoc Mechanistic and Robustness Analyses}\label{sec:posthoc}
Following the locked Icentia11k confirmatory evaluation, additional descriptive analyses were conducted using the locked models, fixed hyperparameters and saved per-record predictions to characterize clinical operating points and patient-level distributions (Section~\ref{sec:icentia}), F-abstention accounting (Section~\ref{sec:ablation}), output-only and class-weighted BN comparators (Section~\ref{sec:sfda}), sub-annotation and N--V contrast relationships (Section~\ref{sec:discussion}), and patient-specificity decomposition and front-end robustness (Section~\ref{sec:posthoc_results}; Supplementary Tables~3--6). These analyses do not re-test the confirmatory hypothesis family (H1 and the Holm contrasts), and their confidence intervals, where reported, are descriptive patient-cluster bootstrap intervals ($B=10{,}000$) using the same cluster resampling weights. Specifically, Prior-EM applied expectation--maximization to the \arm{G0} posteriors of the prefix beats (up to 100 iterations, convergence tolerance $10^{-8}$) and adjusted the evaluation logits by $\log(\pi_t/\pi_s)$ under a uniform source prior $\pi_s$; LAME used concave--convex updates ($\lambda=1$, up to 100 steps) with a binary 5-nearest-neighbor affinity graph constructed on L2-normalized Log-Euclidean vectorizations of $C(x)$. For the fiducial-perturbation experiments, uniform ($\pm a$) or Gaussian ($\sigma=a$) jitter with $a\in\{10,25,50\}$\,ms was applied to every realigned fiducial position using a single random draw per record and condition shared across seeds. Annotation-free evaluation used the \texttt{xqrs} detector in the WFDB Python package (version 4.3.1; based on classical QRS detection principles\cite{pan1985realtime}) on the resampled 360\,Hz signal, realigned the detected peaks with the fixed rule, and matched detections one-to-one to reference annotations within $\pm150$\,ms in accordance with ANSI/AAMI EC57\cite{aami2012ec57}, counting missed reference beats as class-specific false negatives and unmatched detections predicted as S or V as false positives.

\section*{Data availability}
All databases used in this study are publicly available on PhysioNet: MIT-BIH Arrhythmia Database (\url{https://doi.org/10.13026/C2F305}), St.\ Petersburg INCART 12-lead Arrhythmia Database (\url{https://doi.org/10.13026/C2V88N}), MIT-BIH Supraventricular Arrhythmia Database (\url{https://doi.org/10.13026/C2V30W}) and Icentia11k Single Lead Continuous Raw Electrocardiogram Dataset (\url{https://doi.org/10.13026/kk0v-r952}). The summary metrics, bootstrap contrasts and source data underlying all figures and tables are provided in the main text and Supplementary Information; per-record intermediate outputs and the summary tables underlying Supplementary Tables~3--6 are archived in the study repository alongside the analysis code.

\section*{Code availability}
The analysis code, evaluation-protocol records and figure-generation scripts are archived in the study repository. The complete codebase and reproduction scripts will be made publicly available on GitHub upon publication and are available from the corresponding author upon reasonable request during peer review.

\section*{Ethics declarations}
This study analyzed only publicly available, de-identified human ECG recordings distributed through PhysioNet under open data licenses (MIT-BIH, INCART, SVDB and Icentia11k); no new data were collected from human participants and no direct patient interaction occurred. In accordance with institutional and international guidelines on secondary analysis of public de-identified datasets, separate institutional review board approval and informed consent were not required.

\section*{Author contributions}
J.K. conceived the study, designed the methodology, implemented the software and experimental pipelines, performed the formal and statistical analyses, created the visualizations, and wrote and revised the manuscript.

\section*{Competing interests}
J.K. serves as Head of AI at HUINNO Co., Ltd. This study was conducted independently using only publicly available datasets (PhysioNet) and does not evaluate any commercial hardware or software product of HUINNO Co., Ltd. The author declares no other competing interests.

\bibliographystyle{unsrtnat}
\bibliography{main}

\begin{thebibliography}{45}
\providecommand{\natexlab}[1]{#1}
\providecommand{\url}[1]{\texttt{#1}}
\expandafter\ifx\csname urlstyle\endcsname\relax
  \providecommand{\doi}[1]{doi: #1}\else
  \providecommand{\doi}{doi: \begingroup \urlstyle{rm}\Url}\fi

\bibitem[Luz et~al.(2016)Luz, Schwartz, C{\'a}mara-Ch{\'a}vez, and Menotti]{luz2016survey}
Eduardo Jos{\'e} da~S. Luz, William~Robson Schwartz, Guillermo C{\'a}mara-Ch{\'a}vez, and David Menotti.
\newblock {ECG}-based heartbeat classification for arrhythmia detection: A survey.
\newblock \emph{Computer Methods and Programs in Biomedicine}, 127:\penalty0 144--164, 2016.
\newblock \doi{10.1016/j.cmpb.2015.12.008}.

\bibitem[de~Chazal et~al.(2004)de~Chazal, O'Dwyer, and Reilly]{dechazal2004automatic}
Philip de~Chazal, Maria O'Dwyer, and Richard~B. Reilly.
\newblock Automatic classification of heartbeats using {ECG} morphology and heartbeat interval features.
\newblock \emph{IEEE Transactions on Biomedical Engineering}, 51\penalty0 (7):\penalty0 1196--1206, 2004.
\newblock \doi{10.1109/TBME.2004.827359}.

\bibitem[Hu et~al.(1997)Hu, Palreddy, and Tompkins]{hu1997patient}
Yu~Hen Hu, S.~Palreddy, and Willis~J. Tompkins.
\newblock A patient-adaptable {ECG} beat classifier using a mixture of experts approach.
\newblock \emph{IEEE Transactions on Biomedical Engineering}, 44\penalty0 (9):\penalty0 891--900, 1997.
\newblock \doi{10.1109/10.623058}.

\bibitem[de~Chazal and Reilly(2006)]{dechazal2006patient}
Philip de~Chazal and Richard~B. Reilly.
\newblock A patient-adapting heartbeat classifier using {ECG} morphology and heartbeat interval features.
\newblock \emph{IEEE Transactions on Biomedical Engineering}, 53\penalty0 (12):\penalty0 2535--2543, 2006.
\newblock \doi{10.1109/tbme.2006.883802}.

\bibitem[Llamedo and Martinez(2012)]{llamedo2012automatic}
M.~Llamedo and J.~P. Martinez.
\newblock An automatic patient-adapted {ECG} heartbeat classifier allowing expert assistance.
\newblock \emph{IEEE Transactions on Biomedical Engineering}, 59\penalty0 (8):\penalty0 2312--2320, 2012.
\newblock \doi{10.1109/tbme.2012.2202662}.

\bibitem[Ince et~al.(2009)Ince, Kiranyaz, and Gabbouj]{ince2009generic}
Turker Ince, Serkan Kiranyaz, and Moncef Gabbouj.
\newblock A generic and robust system for automated patient-specific classification of {ECG} signals.
\newblock \emph{IEEE Transactions on Biomedical Engineering}, 56\penalty0 (5):\penalty0 1415--1426, 2009.
\newblock \doi{10.1109/tbme.2009.2013934}.

\bibitem[Kiranyaz et~al.(2016)Kiranyaz, Ince, and Gabbouj]{kiranyaz2016realtime}
Serkan Kiranyaz, Turker Ince, and Moncef Gabbouj.
\newblock Real-time patient-specific {ECG} classification by {1-D} convolutional neural networks.
\newblock \emph{IEEE Transactions on Biomedical Engineering}, 63\penalty0 (3):\penalty0 664--675, 2016.
\newblock \doi{10.1109/tbme.2015.2468589}.

\bibitem[Hannun et~al.(2019)Hannun, Rajpurkar, Haghpanahi, Tison, Bourn, Turakhia, and Ng]{hannun2019cardiologist}
Awni~Y. Hannun, Pranav Rajpurkar, Masoumeh Haghpanahi, Geoffrey~H. Tison, Codie Bourn, Mintu~P. Turakhia, and Andrew~Y. Ng.
\newblock Cardiologist-level arrhythmia detection and classification in ambulatory electrocardiograms using a deep neural network.
\newblock \emph{Nature Medicine}, 25\penalty0 (1):\penalty0 65--69, 2019.
\newblock \doi{10.1038/s41591-018-0268-3}.

\bibitem[Li et~al.(2018)Li, Wang, Shi, Hou, and Liu]{li2018adabn}
Yanghao Li, Naiyan Wang, Jianping Shi, Xiaodi Hou, and Jiaying Liu.
\newblock Adaptive batch normalization for practical domain adaptation.
\newblock \emph{Pattern Recognition}, 80:\penalty0 109--117, 2018.
\newblock \doi{10.1016/j.patcog.2018.03.005}.

\bibitem[Wang et~al.(2021)Wang, Shelhamer, Liu, Olshausen, and Darrell]{wang2021tent}
Dequan Wang, Evan Shelhamer, Shaoteng Liu, Bruno Olshausen, and Trevor Darrell.
\newblock {Tent}: Fully test-time adaptation by entropy minimization.
\newblock In \emph{International Conference on Learning Representations (ICLR)}, 2021.

\bibitem[Iwasawa and Matsuo(2021)]{iwasawa2021t3a}
Yusuke Iwasawa and Yutaka Matsuo.
\newblock Test-time classifier adjustment module for model-agnostic domain generalization.
\newblock In \emph{Advances in Neural Information Processing Systems (NeurIPS)}, volume~34, pages 2427--2440, 2021.

\bibitem[Schneider et~al.(2020)Schneider, Rusak, Eck, Bringmann, Brendel, and Bethge]{schneider2020covariate}
Steffen Schneider, Evgenia Rusak, Luisa Eck, Oliver Bringmann, Wieland Brendel, and Matthias Bethge.
\newblock Improving robustness against common corruptions by covariate shift adaptation.
\newblock In \emph{Advances in Neural Information Processing Systems (NeurIPS)}, volume~33, pages 11539--11551, 2020.

\bibitem[Liang et~al.(2020)Liang, Hu, and Feng]{liang2020shot}
Jian Liang, Dapeng Hu, and Jiashi Feng.
\newblock Do we really need to access the source data? {Source} hypothesis transfer for unsupervised domain adaptation.
\newblock In \emph{Proceedings of the 37th International Conference on Machine Learning (ICML)}, volume 119, pages 6028--6039, 2020.

\bibitem[Wang et~al.(2022)Wang, Fink, Van~Gool, and Dai]{wang2022cotta}
Qin Wang, Olga Fink, Luc Van~Gool, and Dengxin Dai.
\newblock Continual test-time domain adaptation.
\newblock In \emph{2022 IEEE/CVF Conference on Computer Vision and Pattern Recognition (CVPR)}, pages 7191--7201. IEEE, 2022.
\newblock \doi{10.1109/cvpr52688.2022.00706}.

\bibitem[Gong et~al.(2022)Gong, Jeong, Kim, Kim, Shin, and Lee]{gong2022note}
Taesik Gong, Jongheon Jeong, Taewon Kim, Yewon Kim, Jinwoo Shin, and Sung-Ju Lee.
\newblock {NOTE:} robust continual test-time adaptation against temporal correlation.
\newblock In \emph{Advances in Neural Information Processing Systems 35}, pages 27253--27266. Neural Information Processing Systems Foundation, Inc. (NeurIPS), 2022.
\newblock \doi{10.52202/068431-1976}.

\bibitem[Niu et~al.(2023)Niu, Wu, Zhang, Wen, Chen, Zhao, and Tan]{niu2023sar}
Shuaicheng Niu, Jiaxiang Wu, Yifan Zhang, Zhiquan Wen, Yaofo Chen, Peilin Zhao, and Mingkui Tan.
\newblock Towards stable test-time adaptation in dynamic wild world.
\newblock In \emph{International Conference on Learning Representations (ICLR)}, 2023.

\bibitem[Yuan et~al.(2023)Yuan, Xie, and Li]{yuan2023rotta}
Longhui Yuan, Binhui Xie, and Shuang Li.
\newblock Robust test-time adaptation in dynamic scenarios.
\newblock In \emph{2023 IEEE/CVF Conference on Computer Vision and Pattern Recognition (CVPR)}, pages 15922--15932. IEEE, 2023.
\newblock \doi{10.1109/cvpr52729.2023.01528}.

\bibitem[Saerens et~al.(2002)Saerens, Latinne, and Decaestecker]{saerens2002adjusting}
Marco Saerens, Patrice Latinne, and Christine Decaestecker.
\newblock Adjusting the outputs of a classifier to new a priori probabilities: A simple procedure.
\newblock \emph{Neural Computation}, 14\penalty0 (1):\penalty0 21--41, 2002.
\newblock \doi{10.1162/089976602753284446}.

\bibitem[Lipton et~al.(2018)Lipton, Wang, and Smola]{lipton2018bbse}
Zachary~C. Lipton, Yu-Xiang Wang, and Alex Smola.
\newblock Detecting and correcting for label shift with black box predictors.
\newblock In \emph{Proceedings of the 35th International Conference on Machine Learning (ICML)}, volume~80, pages 3122--3130, 2018.

\bibitem[Alexandari et~al.(2020)Alexandari, Kundaje, and Shrikumar]{alexandari2020bcts}
Amr Alexandari, Anshul Kundaje, and Avanti Shrikumar.
\newblock Maximum likelihood with bias-corrected calibration is hard-to-beat at label shift adaptation.
\newblock In \emph{Proceedings of the 37th International Conference on Machine Learning (ICML)}, volume 119, pages 222--232, 2020.

\bibitem[Boudiaf et~al.(2022)Boudiaf, Mueller, Ayed, and Bertinetto]{boudiaf2022lame}
Malik Boudiaf, Romain Mueller, Ismail~Ben Ayed, and Luca Bertinetto.
\newblock Parameter-free online test-time adaptation.
\newblock In \emph{2022 IEEE/CVF Conference on Computer Vision and Pattern Recognition (CVPR)}, pages 8334--8343. IEEE, 2022.
\newblock \doi{10.1109/cvpr52688.2022.00816}.

\bibitem[Barachant et~al.(2012)Barachant, Bonnet, Congedo, and Jutten]{barachant2012multiclass}
Alexandre Barachant, St{\'e}phane Bonnet, Marco Congedo, and Christian Jutten.
\newblock Multiclass brain--computer interface classification by {Riemannian} geometry.
\newblock \emph{IEEE Transactions on Biomedical Engineering}, 59\penalty0 (4):\penalty0 920--928, 2012.
\newblock \doi{10.1109/tbme.2011.2172210}.

\bibitem[Congedo et~al.(2017)Congedo, Barachant, and Bhatia]{congedo2017primer}
Marco Congedo, Alexandre Barachant, and Rajendra Bhatia.
\newblock {Riemannian} geometry for {EEG}-based brain-computer interfaces; a primer and a review.
\newblock \emph{Brain-Computer Interfaces}, 4\penalty0 (3):\penalty0 155--174, 2017.
\newblock \doi{10.1080/2326263x.2017.1297192}.

\bibitem[Huang and Van~Gool(2017)]{huang2017spdnet}
Zhiwu Huang and Luc Van~Gool.
\newblock A {Riemannian} network for {SPD} matrix learning.
\newblock In \emph{Proceedings of the AAAI Conference on Artificial Intelligence}, volume~31, 2017.
\newblock \doi{10.1609/aaai.v31i1.10866}.

\bibitem[Brooks et~al.(2019)Brooks, Schwander, Barbaresco, Schneider, and Cord]{brooks2019riemannian}
Daniel Brooks, Olivier Schwander, Fr{\'e}d{\'e}ric Barbaresco, Jean-Yves Schneider, and Matthieu Cord.
\newblock {Riemannian} batch normalization for {SPD} neural networks.
\newblock In \emph{Advances in Neural Information Processing Systems (NeurIPS)}, volume~32, 2019.

\bibitem[Kobler et~al.(2022)Kobler, Hirayama, Zhao, and Kawanabe]{kobler2022spdbn}
Reinmar Kobler, Jun-Ichiro Hirayama, Qibin Zhao, and Motoaki Kawanabe.
\newblock {SPD} domain-specific batch normalization to crack interpretable unsupervised domain adaptation in {EEG}.
\newblock In \emph{Advances in Neural Information Processing Systems 35}, pages 6219--6235. Neural Information Processing Systems Foundation, Inc. (NeurIPS), 2022.
\newblock \doi{10.52202/068431-0450}.

\bibitem[Yair et~al.(2019)Yair, Ben-Chen, and Talmon]{yair2019parallel}
Or~Yair, Mirela Ben-Chen, and Ronen Talmon.
\newblock Parallel transport on the cone manifold of {SPD} matrices for domain adaptation.
\newblock \emph{IEEE Transactions on Signal Processing}, 67\penalty0 (7):\penalty0 1797--1811, 2019.
\newblock \doi{10.1109/tsp.2019.2894801}.

\bibitem[Courty et~al.(2017)Courty, Flamary, Tuia, and Rakotomamonjy]{courty2017optimal}
Nicolas Courty, R{\'e}mi Flamary, Devis Tuia, and Alain Rakotomamonjy.
\newblock Optimal transport for domain adaptation.
\newblock \emph{IEEE Transactions on Pattern Analysis and Machine Intelligence}, 39\penalty0 (9):\penalty0 1853--1865, 2017.
\newblock \doi{10.1109/tpami.2016.2615921}.

\bibitem[Zanini et~al.(2018)Zanini, Congedo, Jutten, Said, and Berthoumieu]{zanini2018transfer}
Paolo Zanini, Marco Congedo, Christian Jutten, Salem Said, and Yannick Berthoumieu.
\newblock Transfer learning: A {Riemannian} geometry framework with applications to brain--computer interfaces.
\newblock \emph{IEEE Transactions on Biomedical Engineering}, 65\penalty0 (5):\penalty0 1107--1116, 2018.
\newblock \doi{10.1109/tbme.2017.2742541}.

\bibitem[Rodrigues et~al.(2019)Rodrigues, Jutten, and Congedo]{rodrigues2019rpa}
Pedro Luiz~Coelho Rodrigues, Christian Jutten, and Marco Congedo.
\newblock {Riemannian} {Procrustes} analysis: Transfer learning for brain--computer interfaces.
\newblock \emph{IEEE Transactions on Biomedical Engineering}, 66\penalty0 (8):\penalty0 2390--2401, 2019.
\newblock \doi{10.1109/tbme.2018.2889705}.

\bibitem[Moody and Mark(2001)]{moody2001impact}
G.B. Moody and R.G. Mark.
\newblock The impact of the {MIT-BIH} arrhythmia database.
\newblock \emph{IEEE Engineering in Medicine and Biology Magazine}, 20\penalty0 (3):\penalty0 45--50, 2001.
\newblock \doi{10.1109/51.932724}.

\bibitem[Goldberger et~al.(2000)Goldberger, Amaral, Glass, Hausdorff, Ivanov, Mark, Mietus, Moody, Peng, and Stanley]{goldberger2000physionet}
Ary~L. Goldberger, Luis A.~N. Amaral, Leon Glass, Jeffrey~M. Hausdorff, Plamen~Ch. Ivanov, Roger~G. Mark, Joseph~E. Mietus, George~B. Moody, Chung-Kang Peng, and H.~Eugene Stanley.
\newblock {PhysioBank}, {PhysioToolkit}, and {PhysioNet}: Components of a new research resource for complex physiologic signals.
\newblock \emph{Circulation}, 101\penalty0 (23):\penalty0 e215--e220, 2000.
\newblock \doi{10.1161/01.CIR.101.23.e215}.

\bibitem[AAMI(2013)]{aami2012ec57}
\emph{{ANSI/AAMI EC57:2012/(R)2020}: Testing and reporting performance results of cardiac rhythm and {ST} segment measurement algorithms}.
\newblock Association for the Advancement of Medical Instrumentation, Arlington, VA, USA, 2013.

\bibitem[Pennec et~al.(2006)Pennec, Fillard, and Ayache]{pennec2006riemannian}
Xavier Pennec, Pierre Fillard, and Nicholas Ayache.
\newblock A {Riemannian} framework for tensor computing.
\newblock \emph{International Journal of Computer Vision}, 66\penalty0 (1):\penalty0 41--66, 2006.
\newblock \doi{10.1007/s11263-005-3222-z}.

\bibitem[Arsigny et~al.(2007)Arsigny, Fillard, Pennec, and Ayache]{arsigny2007geometric}
Vincent Arsigny, Pierre Fillard, Xavier Pennec, and Nicholas Ayache.
\newblock Geometric means in a novel vector space structure on symmetric positive-definite matrices.
\newblock \emph{SIAM Journal on Matrix Analysis and Applications}, 29\penalty0 (1):\penalty0 328--347, 2007.
\newblock \doi{10.1137/050637996}.

\bibitem[Bhatia(2009)]{bhatia2009positive}
Rajendra Bhatia.
\newblock \emph{{Positive Definite Matrices}}.
\newblock Princeton University Press, 2009.
\newblock \doi{10.2307/j.ctt7rxv2}.

\bibitem[Moakher(2005)]{moakher2005geometric}
Maher Moakher.
\newblock A differential geometric approach to the geometric mean of symmetric positive-definite matrices.
\newblock \emph{SIAM Journal on Matrix Analysis and Applications}, 26\penalty0 (3):\penalty0 735--747, 2005.
\newblock \doi{10.1137/s0895479803436937}.

\bibitem[Dowson and Landau(1982)]{dowson1982frechet}
D.~C. Dowson and B.~V. Landau.
\newblock The {Fr{\'e}chet} distance between multivariate normal distributions.
\newblock \emph{Journal of Multivariate Analysis}, 12\penalty0 (3):\penalty0 450--455, 1982.
\newblock \doi{10.1016/0047-259x(82)90077-x}.

\bibitem[Olkin and Pukelsheim(1982)]{olkin1982distance}
I.~Olkin and F.~Pukelsheim.
\newblock The distance between two random vectors with given dispersion matrices.
\newblock \emph{Linear Algebra and its Applications}, 48:\penalty0 257--263, 1982.
\newblock \doi{10.1016/0024-3795(82)90112-4}.

\bibitem[Bhatia et~al.(2019)Bhatia, Jain, and Lim]{bhatia2019bures}
Rajendra Bhatia, Tanvi Jain, and Yongdo Lim.
\newblock On the {Bures--Wasserstein} distance between positive definite matrices.
\newblock \emph{Expositiones Mathematicae}, 37\penalty0 (2):\penalty0 165--191, 2019.
\newblock \doi{10.1016/j.exmath.2018.01.002}.

\bibitem[Tihonenko et~al.(2007)Tihonenko, Khaustov, Ivanov, and Rivin]{tihonenko2008incart}
Viktor Tihonenko, Alexander Khaustov, Sergey Ivanov, and Alexei Rivin.
\newblock {St.\ Petersburg} {Institute} of {Cardiological} {Technics} 12-lead arrhythmia database.
\newblock PhysioNet, 2007.

\bibitem[Greenwald(1990)]{greenwald1990svdb}
Scott~David Greenwald.
\newblock \emph{Improved detection and classification of arrhythmias in noise-corrupted electrocardiograms using contextual information}.
\newblock PhD thesis, Harvard--MIT Division of Health Sciences and Technology, Cambridge, MA, USA, 1990.
\newblock Source of the MIT-BIH Supraventricular Arrhythmia Database (PhysioNet, doi:10.13026/C2V30W).

\bibitem[Tan et~al.(2022)Tan, Ortiz-Gagn{\'e}, Beaudoin-Gagnon, Fecteau, Courville, Bengio, and Cohen]{tan2022icentia}
Shawn Tan, Satya Ortiz-Gagn{\'e}, Nicolas Beaudoin-Gagnon, Pierre Fecteau, Aaron Courville, Yoshua Bengio, and Joseph~Paul Cohen.
\newblock {Icentia11k} single lead continuous raw electrocardiogram dataset.
\newblock PhysioNet, 2022.

\bibitem[Tan et~al.(2019)Tan, Androz, Chamseddine, Fecteau, Courville, Bengio, and Cohen]{tan2021icentia}
Shawn Tan, Guillaume Androz, Ahmad Chamseddine, Pierre Fecteau, Aaron Courville, Yoshua Bengio, and Joseph~Paul Cohen.
\newblock {Icentia11k}: An unsupervised representation learning dataset for arrhythmia subtype discovery.
\newblock arXiv:1910.09570, 2019.

\bibitem[Pan and Tompkins(1985)]{pan1985realtime}
Jiapu Pan and Willis~J. Tompkins.
\newblock A real-time {QRS} detection algorithm.
\newblock \emph{IEEE Transactions on Biomedical Engineering}, BME-32\penalty0 (3):\penalty0 230--236, 1985.
\newblock \doi{10.1109/tbme.1985.325532}.

\end{thebibliography}

\clearpage
\newgeometry{margin=2.0cm}\singlespacing
\setcounter{table}{0}
\renewcommand{\tablename}{Supplementary Table}
\renewcommand{\theHtable}{S\arabic{table}}
\phantomsection\addcontentsline{toc}{section}{Supplementary Information}
\begin{center}
{\LARGE Supplementary Information}\\[6pt]
{\large Transport What Shares the Conduction Pathway: Normal-Anchored SPD Geometric Transport\\for Source-Free Patient Adaptation of Holter and Patch ECG}\\[8pt]
JinKook Kim\\[2pt]
{\small AI Division, HUINNO Co., Ltd., Seoul, Republic of Korea; E-mail: jinkook@huinno.com}
\end{center}
\vspace{6pt}
\section*{Supplementary Note 1: Model development stages and confirmatory evaluation}
\hyperref[tab:history]{Supplementary Table~1} summarizes the model development stages, the intermediate negative findings and the final locked confirmatory evaluation. The development cohorts (MIT-BIH dev7, DS1-CV and DS2, INCART and SVDB) were used to develop the architecture, the anchor selection, the F buffer and the rate-invariant relA head; the held-out Icentia11k cohort was evaluated once, with all model weights, hyperparameters and decision rules locked.
\hyperref[tab:full]{Supplementary Table~2} lists every metric with its 95\% patient-cluster bootstrap CI for every dataset and cohort (INCART and SVDB with both sources).

\begin{table}[!h]
\centering
\caption{Summary of model development stages, intermediate negative findings, and final locked confirmatory evaluation.}
\label{tab:history}
\scriptsize
\setlength{\tabcolsep}{3pt}
\begin{adjustbox}{max width=\textwidth}
\begin{tabular}{l>{\raggedright\arraybackslash}p{3.0cm}>{\raggedright\arraybackslash}p{4.6cm}>{\raggedright\arraybackslash}p{6.2cm}>{\raggedright\arraybackslash}p{2.0cm}}
\toprule
Stage & Cohort / Split & Model \& Design Evaluated & Key Quantitative Finding & Role \\
\midrule
1 & dev7, oracle anchors & proof of concept and staged tuning & frozen recipe GA 85.85 vs GN 74.68 / G0 78.99 & Development \\
2 & dev7 seeds 0--2 & label-free U1 anchors (selection rules fixed in advance) & GA 70.96 $\pm$ 13.52 did not beat GN 74.69; V collapse identified post hoc & Development (negative) \\
3 & dev7 seeds 3--5 (new) & GA\_SV, locked evaluation & GA\_SV 84.89 $>$ GN 72.11, GA\_All 65.74, G0 62.37 & Development \\
4 & MIT-BIH DS2 & GA\_SV versus GN with Track A and Track B sources & H\_P not confirmed: GA\_SV $-$ GN +0.28 [$-$10.43, +8.67] (Track A, P1) & Development (negative) \\
5 & DS1-CV selection, DS2 post hoc & GA\_NS selected on DS1 only & DS2 GA\_NS 56.87 (A) / 55.88 (B) & Development (post hoc) \\
6 & DS2 post hoc & rhythm stratification; rr8 head not adopted & non-AF-17 GA\_NS $-$ GN +10.03 [+1.51, +14.89] & Development (post hoc) \\
7 & INCART primary, SVDB secondary & original four-dimensional RR head; fiducial realignment; duplicate-annotation removal & H1 not confirmed (+3.92 [$-$0.27, +8.46]); H2/H3 untested by the hierarchy; mechanism reproduced & Development (negative) \\
\bottomrule
\end{tabular}
\end{adjustbox}
\end{table}

\clearpage
\begin{landscape}
{\scriptsize\setlength{\tabcolsep}{3pt}
\begin{longtable}{lllllllll}
\caption{Full benchmark (supplementary): all metrics with 95\% patient-cluster bootstrap CI (\%).}\label{tab:full}\\
\toprule
Method & J & S-F1 & S-Se & S-PPV & V-F1 & V-Se & N-Se & $\Delta J$ vs GA\_NS \\
\midrule
\endfirsthead
\caption[]{Full benchmark (continued).}\\
\toprule
Method & J & S-F1 & S-Se & S-PPV & V-F1 & V-Se & N-Se & $\Delta J$ vs GA\_NS \\
\midrule
\endhead
\midrule
\multicolumn{9}{r}{\textit{continued on next page}} \\
\endfoot
\bottomrule
\endlastfoot
\multicolumn{9}{l}{\textit{Icentia11k (MIT-BIH-44 source) --- all-comers (973 records, 493 clusters)}} \\
G0 (no adaptation) & 34.5 [27.4, 41.8] & 25.0 [18.1, 32.0] & 61.0 [52.5, 69.2] & 16.6 [11.2, 22.9] & 44.1 [31.6, 56.4] & 75.5 [68.6, 81.1] & 73.0 [70.5, 75.5] & +14.1 [+8.6, +19.3] \\
AdaBN & 24.7 [19.7, 29.9] & 24.9 [18.7, 31.0] & 34.1 [27.3, 41.8] & 20.4 [13.9, 28.0] & 24.6 [17.0, 32.7] & 92.0 [87.6, 95.6] & 66.6 [63.9, 69.1] & +23.9 [+19.6, +28.2] \\
Tent & 28.3 [22.8, 33.9] & 30.3 [23.2, 37.2] & 38.0 [30.4, 46.5] & 25.9 [17.9, 35.1] & 26.2 [18.2, 34.9] & 91.7 [87.2, 95.2] & 68.7 [65.7, 71.6] & +20.4 [+16.0, +24.5] \\
T3A & 25.6 [19.7, 31.3] & 10.0 [6.8, 13.5] & 46.2 [39.6, 52.7] & 5.6 [3.7, 7.9] & 41.2 [29.9, 52.1] & 77.5 [70.8, 83.4] & 74.3 [72.4, 76.1] & +23.1 [+18.3, +27.7] \\
RPA-Recenter & 42.8 [35.8, 49.0] & 31.0 [24.1, 37.9] & 56.7 [49.6, 65.1] & 21.3 [15.2, 28.5] & 54.6 [42.9, 64.6] & 74.5 [67.5, 81.0] & 94.6 [93.8, 95.3] & +5.9 [+2.8, +9.1] \\
GN & 35.2 [30.0, 39.5] & 6.8 [4.5, 9.6] & 18.4 [13.2, 25.0] & 4.3 [2.7, 6.3] & 63.5 [53.4, 71.7] & 69.9 [62.3, 76.7] & 94.1 [93.3, 94.9] & +13.5 [+9.4, +17.8] \\
GA\_All & 44.7 [37.8, 50.9] & 31.8 [24.6, 39.2] & 61.1 [53.2, 70.4] & 21.5 [15.4, 28.8] & 57.5 [46.1, 67.1] & 72.7 [65.3, 79.4] & 94.9 [94.2, 95.6] & +4.0 [+1.2, +6.9] \\
\end{longtable}
}
\end{landscape}

\section*{Supplementary Note 2: Post hoc mechanistic and robustness analyses}
Supplementary Tables~3--6 present descriptive post hoc analyses conducted after the locked Icentia11k confirmatory evaluation using the locked models, fixed hyperparameters and saved per-record predictions (see Section~\ref{sec:posthoc}). These analyses do not re-test the primary hypothesis H1 or the Holm multiple-comparison family. Pooled metrics follow the main-text protocol (beat counts summed across records within each seed, metrics computed per seed and averaged across the six seeds), with Icentia11k evaluated under the primary F-offset rule ($\delta_F=0.5$). Reported confidence intervals are descriptive patient-cluster bootstrap intervals ($B=10{,}000$) using the locked cluster resampling weights.
\begin{itemize}
\item Supplementary Table~3: complete $4\times4$ confusion matrices for every evaluation set and method (3a); standard four-class N-Se versus effective non-alarm specificity (3b); prevalence-normalized likelihood ratios, three-class balanced accuracy and false alarms per hour (3c); patient- and record-level $\Delta J$ distributions (3d); and record-level PVC and PAC burden agreement (3e).
\item Supplementary Table~4: decomposition of the adaptation gain into dataset-level and patient-specific components alongside null control maps (4a, 4b); MIT-BIH sub-annotation stratification of prototype distance ratios (4c); and Icentia11k baseline N--V morphological contrast versus V prototype repulsion (4d).
\item Supplementary Table~5: output-only label-shift and transductive comparators (Prior-EM and LAME) across all evaluation sets (5a); Prior-EM convergence and minority-class prior collapse (5b); oracle class-weighted AdaBN diagnostic (5c); and pooled $J$ stratified by the natural prefix normal-beat fraction (5d).
\item Supplementary Table~6: sensitivity to admitting unclassified (Q) beats into the Icentia11k adaptation prefix (6a, 6b); robustness to fiducial timing jitter (6c); and end-to-end annotation-free evaluation using the \texttt{xqrs} R-peak detector (6d, 6e).
\end{itemize}

\clearpage
\section*{Supplementary Table 3: Confusion matrices and clinical operating characteristics}

\renewcommand{\thetable}{3a}\renewcommand{\theHtable}{S3a}
{\scriptsize\setlength{\tabcolsep}{2pt}
\setlength{\LTpost}{2pt}
\begin{longtable}{llrrrr}
\caption{Complete $4\times4$ confusion matrices (6-seed mean pooled beat counts per seed; row percentage of the true class in parentheses). Rows: true class; columns: predicted class. Icentia11k (primary rule, $\delta_F=0.5$) and the development sets of Table~1 (final configuration). True classes with no beat are omitted.}\label{tab:s3a}\\
\toprule
Model & True & pred.\ N & pred.\ S & pred.\ V & pred.\ F \\
\midrule
\endfirsthead
\multicolumn{6}{l}{\textit{(continued)}}\\
\toprule
Model & True & pred.\ N & pred.\ S & pred.\ V & pred.\ F \\
\midrule
\endhead
\bottomrule
\endlastfoot
\multicolumn{6}{l}{\textbf{Icentia11k all} (true beats per seed: N 3,565,024, S 23,076, V 27,288, F 0)}\\
G0 & N & 2,604,228.8 (73.05) & 81,240.8 (2.28) & 43,279.8 (1.21) & 836,274.5 (23.46) \\
 & S & 3,836.8 (16.63) & 14,081.7 (61.02) & 4,277.7 (18.54) & 879.8 (3.81) \\
 & V & 872.8 (3.20) & 4,839.5 (17.73) & 20,602.5 (75.50) & 973.2 (3.57) \\
AdaBN & N & 2,373,545.5 (66.58) & 31,224.0 (0.88) & 154,023.7 (4.32) & 1,006,230.8 (28.23) \\
 & S & 3,852.3 (16.69) & 7,877.2 (34.14) & 10,870.3 (47.11) & 476.2 (2.06) \\
 & V & 547.2 (2.01) & 754.5 (2.76) & 25,110.7 (92.02) & 875.7 (3.21) \\
Tent & N & 2,449,315.5 (68.70) & 25,027.7 (0.70) & 144,170.2 (4.04) & 946,510.7 (26.55) \\
\end{longtable}
\par\noindent{\footnotesize Counts are pooled across records per seed and averaged over the six seeds under the final locked configuration.\par}}

\renewcommand{\thetable}{3b}\renewcommand{\theHtable}{S3b}
{\scriptsize\setlength{\tabcolsep}{3pt}
\begin{longtable}{lccccccc}
\caption{Two-way reporting of the F output: standard 4-class N-Se (N$\to$N / true N) and effective non-alarm specificity ((N$\to$N + N$\to$F) / true N), in \%. Merging the predicted-F column into N leaves all S and V counts unchanged, so S-Se, S-PPV, V-Se, V-PPV and $J$ are identical under both readings.}\label{tab:s3b}\\
\toprule
Model & Icentia11k all & Icentia11k non-AF & DS1-CV & DS2 A & DS2 B & INCART S44 & SVDB S44 \\
\midrule
\endfirsthead
\multicolumn{8}{l}{\textit{(continued)}}\\
\toprule
Model & Icentia11k all & Icentia11k non-AF & DS1-CV & DS2 A & DS2 B & INCART S44 & SVDB S44 \\
\midrule
\endhead
\bottomrule
\endlastfoot
G0 & 73.05 / 96.51 & 74.17 / 97.78 & 75.45 / 87.79 & 80.82 / 93.24 & 81.95 / 91.23 & 81.04 / 96.32 & 54.65 / 91.84 \\
AdaBN & 66.58 / 94.80 & 67.85 / 96.38 & 67.47 / 83.27 & 68.98 / 89.01 & 66.60 / 86.05 & 72.62 / 95.35 & 54.65 / 92.98 \\
Tent & 68.70 / 95.25 & 69.89 / 96.81 & 71.47 / 85.63 & 74.21 / 91.51 & 72.85 / 89.07 & 76.53 / 95.99 & 55.14 / 93.07 \\
T3A & 74.28 / 93.65 & 75.49 / 94.91 & 66.88 / 84.41 & 73.74 / 86.18 & 72.86 / 85.89 & 71.33 / 90.42 & 60.72 / 94.31 \\
RPA-Recenter & 94.59 / 98.14 & 95.64 / 99.17 & 95.26 / 96.75 & 95.98 / 97.29 & 93.93 / 95.26 & 94.19 / 97.69 & 96.62 / 97.95 \\
GN & 94.14 / 97.04 & 94.44 / 97.28 & 93.15 / 95.05 & 94.37 / 97.43 & 94.48 / 96.55 & 87.57 / 94.48 & 93.26 / 97.24 \\
GA\_All & 94.94 / 98.18 & 95.99 / 99.19 & 94.55 / 96.15 & 96.02 / 97.27 & 93.69 / 95.13 & 92.75 / 97.21 & 96.31 / 97.89 \\
GA\_SV & 93.84 / 96.79 & 94.88 / 97.79 & 93.58 / 95.22 & 95.43 / 96.62 & 93.14 / 94.52 & 90.59 / 94.19 & 94.73 / 96.03 \\
GA\_NS & 93.87 / 98.26 & 94.85 / 99.23 & 93.61 / 96.09 & 94.14 / 97.50 & 92.71 / 95.22 & 88.16 / 97.63 & 92.76 / 97.45 \\
GA\_NS\_Oracle & 94.83 / 98.36 & 95.72 / 99.27 & 94.74 / 96.52 & 95.27 / 97.67 & 93.95 / 95.56 & 91.17 / 97.91 & 94.10 / 97.65 \\
\end{longtable}}

\begin{landscape}
\renewcommand{\thetable}{3c}\renewcommand{\theHtable}{S3c}
{\scriptsize\setlength{\tabcolsep}{3pt}
\begin{longtable}{lrrrrrrcrc}
\caption{Prevalence-normalized operating characteristics on Icentia11k. One-vs-rest LR+ $=$ Se/(1$-$Sp) and LR$-$ $=$ (1$-$Se)/Sp per class (per seed, then averaged); BA3: mean of N-, S- and V-sensitivity (\%). False alarms per hour of evaluated recording (record duration minus the 5-min prefix): V FA $=$ N$\to$V, S+V FA $=$ N$\to$S $+$ N$\to$V; pooled over all records and per-record median [IQR] (6-seed mean counts).}\label{tab:s3c}\\
\toprule
Model & S LR+ & S LR$-$ & V LR+ & V LR$-$ & BA3 & V FA/h & V FA/h rec.\ median [IQR] & S+V FA/h & S+V FA/h rec.\ median [IQR] \\
\midrule
\endfirsthead
\multicolumn{10}{l}{\textit{(continued)}}\\
\toprule
Model & S LR+ & S LR$-$ & V LR+ & V LR$-$ & BA3 & V FA/h & V FA/h rec.\ median [IQR] & S+V FA/h & S+V FA/h rec.\ median [IQR] \\
\midrule
\endhead
\bottomrule
\endlastfoot
\multicolumn{10}{l}{\textbf{Icentia11k all} (total evaluation time 1,052.5\,h)}\\
G0 & 31.6 & 0.398 & 65.3 & 0.248 & 69.86 & 41.12 & 1.54 [0.31, 6.32] & 118.30 & 25.11 [7.55, 84.12] \\
AdaBN & 40.1 & 0.664 & 22.6 & 0.084 & 64.25 & 146.33 & 23.88 [5.24, 107.85] & 176.00 & 38.98 [11.86, 143.90] \\
Tent & 54.9 & 0.624 & 25.1 & 0.087 & 66.12 & 136.97 & 12.63 [2.62, 61.78] & 160.75 & 19.88 [5.24, 87.51] \\
T3A & 9.3 & 0.566 & 53.9 & 0.228 & 65.99 & 48.63 & 2.16 [0.31, 13.10] & 215.15 & 73.03 [17.10, 234.96] \\
RPA-Recenter & 42.2 & 0.439 & 103.3 & 0.257 & 75.28 & 21.39 & 2.16 [0.46, 8.63] & 62.91 & 6.93 [1.85, 25.42] \\
GN & 6.9 & 0.838 & 201.7 & 0.302 & 60.83 & 10.96 & 0.77 [0.00, 3.54] & 100.38 & 18.33 [3.54, 72.26] \\
GA\_All & 42.8 & 0.394 & 127.7 & 0.275 & 76.25 & 17.58 & 1.54 [0.31, 6.93] & 61.58 & 6.32 [1.69, 24.34] \\
\end{longtable}}
\end{landscape}

\begin{landscape}
\renewcommand{\thetable}{3d}\renewcommand{\theHtable}{S3d}
{\scriptsize\setlength{\tabcolsep}{3pt}
\begin{longtable}{llrrrrrrcc}
\caption{Distribution of $\Delta J=J(\text{GA\_NS})-J(\text{G0})$ (pp) over Icentia11k patients (493) and records (973), primary rule. Per unit: $J$ per seed from the unit's pooled counts, then the seed mean. ``strict'': paper definition (undefined if S or V is absent and never predicted); ``available'': mean of the defined F1 values. Win/tie/loss: $\Delta J>0$, $=0$, $<0$ (exact), and with a $\pm0.5$\,pp tie band.}\label{tab:s3d}\\
\toprule
Level & $J$ & evaluable & P10 & P25 & P50 & P75 & P90 & W/T/L (\%) & W/T/L ($\pm$0.5\,pp) \\
\midrule
\endfirsthead
\multicolumn{10}{l}{\textit{(continued)}}\\
\toprule
Level & $J$ & evaluable & P10 & P25 & P50 & P75 & P90 & W/T/L (\%) & W/T/L ($\pm$0.5\,pp) \\
\midrule
\endhead
\bottomrule
\endlastfoot
record & strict & 751/973 & $-$5.56 & $+$0.00 & $+$0.70 & $+$11.53 & $+$23.55 & 388/203/160 (51.7/27.0/21.3) & 379/227/145 \\
record & available & 945/973 & $-$6.05 & $+$0.00 & $+$0.00 & $+$14.52 & $+$31.33 & 460/308/177 (48.7/32.6/18.7) & 451/330/164 \\
patient & strict & 451/493 & $-$4.90 & $+$0.00 & $+$2.69 & $+$13.01 & $+$23.15 & 272/92/87 (60.3/20.4/19.3) & 268/100/83 \\
patient & available & 489/493 & $-$5.42 & $+$0.00 & $+$2.27 & $+$14.01 & $+$25.58 & 282/113/94 (57.7/23.1/19.2) & 278/121/90 \\
\end{longtable}}
\end{landscape}

\renewcommand{\thetable}{3e}\renewcommand{\theHtable}{S3e}
{\scriptsize\setlength{\tabcolsep}{3pt}
\begin{longtable}{llrrrrrr}
\caption{Record-level burden on Icentia11k (973 records): true V (PVC) and S (PAC) burden (\% of evaluated beats) versus predicted burden (6-seed mean), Pearson $r$, Spearman $\rho$, mean and median absolute error (pp).}\label{tab:s3e}\\
\toprule
Class & Model & true (mean \%) & predicted (mean \%) & $r$ & $\rho$ & MAE & median AE \\
\midrule
\endfirsthead
\multicolumn{8}{l}{\textit{(continued)}}\\
\toprule
Class & Model & true (mean \%) & predicted (mean \%) & $r$ & $\rho$ & MAE & median AE \\
\midrule
\endhead
\bottomrule
\endlastfoot
V & G0 & 0.883 & 1.827 & 0.447 & 0.587 & 1.378 & 0.080 \\
V & AdaBN & 0.883 & 5.357 & 0.307 & 0.365 & 4.706 & 0.946 \\
V & Tent & 0.883 & 5.025 & 0.289 & 0.410 & 4.382 & 0.520 \\
V & T3A & 0.883 & 2.141 & 0.487 & 0.533 & 1.626 & 0.094 \\
V & RPA-Recenter & 0.883 & 1.313 & 0.710 & 0.564 & 0.867 & 0.109 \\
V & GN & 0.883 & 0.965 & 0.731 & 0.634 & 0.621 & 0.044 \\
V & GA\_All & 0.883 & 1.167 & 0.741 & 0.570 & 0.759 & 0.087 \\
V & GA\_SV & 0.883 & 2.542 & 0.409 & 0.432 & 2.085 & 0.249 \\
V & GA\_NS & 0.883 & 1.073 & 0.700 & 0.627 & 0.675 & 0.053 \\
V & GA\_NS\_Oracle & 0.883 & 1.023 & 0.723 & 0.635 & 0.642 & 0.050 \\
\midrule
S & G0 & 0.665 & 2.859 & 0.401 & 0.164 & 2.430 & 0.688 \\
S & AdaBN & 0.665 & 1.233 & 0.425 & 0.226 & 1.168 & 0.205 \\
S & Tent & 0.665 & 1.097 & 0.448 & 0.335 & 1.013 & 0.096 \\
S & T3A & 0.665 & 5.186 & 0.191 & 0.044 & 4.834 & 1.894 \\
S & RPA-Recenter & 0.665 & 1.794 & 0.430 & 0.378 & 1.457 & 0.127 \\
S & GN & 0.665 & 2.831 & 0.070 & 0.051 & 2.950 & 0.617 \\
S & GA\_All & 0.665 & 1.894 & 0.433 & 0.378 & 1.525 & 0.133 \\
S & GA\_SV & 0.665 & 1.938 & 0.439 & 0.399 & 1.514 & 0.109 \\
S & GA\_NS & 0.665 & 1.942 & 0.440 & 0.397 & 1.519 & 0.109 \\
S & GA\_NS\_Oracle & 0.665 & 1.902 & 0.450 & 0.415 & 1.478 & 0.102 \\
\end{longtable}}

\clearpage
\section*{Supplementary Table 4: Decomposition of the adaptation gain, sub-annotation stratification and N--V contrast}

\renewcommand{\thetable}{4a}\renewcommand{\theHtable}{S4a}
{\scriptsize\setlength{\tabcolsep}{3pt}
\begin{longtable}{lrcr}
\caption{Decomposition of the adaptation gain without retraining. $A_{\mathrm{device}}$: leave-one-patient-out dataset-global map $T(P_N\to M_{\mathrm{dev}})$, $M_{\mathrm{dev}}$ the AIRM mean of the U1 anchor means $\hat{M}_N$ of all other patients in the same evaluation unit (applied to every record; ``anchored-only'': identity where the record has no U1 anchors). $A_{\mathrm{shuffled}}$: map of a randomly drawn record of another patient. Random tangent: $P_N^{1/2}\exp(rU)P_N^{1/2}$ with $U$ a random symmetric unit-Frobenius direction and $r=\|\log(P_N^{-1/2}\hat{M}_NP_N^{-1/2})\|_F$ of the record's own map. $\Delta J$ in pp; 95\% patient-cluster bootstrap intervals (frozen weights, $B=10{,}000$; descriptive). Share: \% of the total GA\_NS$-$G0 gain.}\label{tab:s4a}\\
\toprule
Component & $\Delta J$ & 95\% CI & share (\%) \\
\midrule
\endfirsthead
\multicolumn{4}{l}{\textit{(continued)}}\\
\toprule
Component & $\Delta J$ & 95\% CI & share (\%) \\
\midrule
\endhead
\bottomrule
\endlastfoot
\multicolumn{4}{l}{\textbf{Icentia11k all} (973 records, 493 patients)}\\
total (GA\_NS - G0) & $+$14.12 & [$+$8.62, $+$19.25] & 100.0 \\
dataset/lead ($A_{\mathrm{device}}$ - G0) & $+$0.50 & [$-$1.59, $+$2.60] & 3.5 \\
patient-specific (GA\_NS - $A_{\mathrm{device}}$) & $+$13.62 & [$+$8.21, $+$18.59] & 96.5 \\
dataset/lead, anchored-only variant & $+$0.53 & [$-$1.52, $+$2.62] & 3.8 \\
patient-specific, anchored-only variant & $+$13.58 & [$+$8.17, $+$18.53] & 96.2 \\
$A_{\mathrm{shuffled}}$ - G0 & $-$4.74 & [$-$7.65, $-$2.11] & -33.6 \\
GA\_NS - $A_{\mathrm{shuffled}}$ & $+$18.85 & [$+$14.65, $+$22.99] & 133.6 \\
\end{longtable}}

\begin{landscape}
\renewcommand{\thetable}{4b}\renewcommand{\theHtable}{S4b}
{\scriptsize\setlength{\tabcolsep}{3pt}
\begin{longtable}{llcrcrcr}
\caption{Distance ratios $d_{\mathrm{LE}}(AP_cA^{\top},\bar{C}_c)/d_{\mathrm{LE}}(P_c,\bar{C}_c)$ under the own map and the three null/decomposition maps: median [IQR] and \% of record\,$\times$\,seed units with ratio $<1$ (classes with $\geq10$ evaluated beats).}\label{tab:s4b}\\
\toprule
Set & Map & N ratio & N $<$1 (\%) & S ratio & S $<$1 (\%) & V ratio & V $<$1 (\%) \\
\midrule
\endfirsthead
\multicolumn{8}{l}{\textit{(continued)}}\\
\toprule
Set & Map & N ratio & N $<$1 (\%) & S ratio & S $<$1 (\%) & V ratio & V $<$1 (\%) \\
\midrule
\endhead
\bottomrule
\endlastfoot
Icentia11k all & own $A_p$ & 0.290 [0.187, 0.473] & 94.1 & 0.560 [0.376, 0.807] & 84.6 & 1.281 [1.002, 1.615] & 21.3 \\
 & $A_{\mathrm{device}}$ & 0.753 [0.662, 0.902] & 84.8 & 0.810 [0.700, 1.019] & 72.8 & 0.999 [0.853, 1.238] & 50.2 \\
 & $A_{\mathrm{shuffled}}$ & 1.114 [0.934, 1.382] & 32.7 & 1.180 [0.975, 1.460] & 27.9 & 1.359 [1.083, 1.736] & 15.1 \\
 & random tangent & 1.434 [1.326, 1.541] & 0.2 & 1.476 [1.353, 1.601] & 0.3 & 1.517 [1.328, 1.757] & 0.5 \\
\midrule
DS2 A & own $A_p$ & 0.249 [0.157, 0.359] & 100.0 & 0.755 [0.634, 1.013] & 75.0 & 1.731 [1.519, 1.880] & 0.0 \\
 & $A_{\mathrm{device}}$ & 0.967 [0.913, 1.036] & 59.8 & 0.963 [0.929, 1.046] & 66.7 & 1.169 [1.047, 1.229] & 11.7 \\
 & $A_{\mathrm{shuffled}}$ & 1.427 [1.206, 1.602] & 7.6 & 1.391 [1.213, 1.505] & 10.4 & 1.742 [1.458, 1.977] & 0.0 \\
 & random tangent & 1.455 [1.336, 1.529] & 0.0 & 1.423 [1.260, 1.550] & 0.0 & 1.763 [1.541, 1.959] & 0.0 \\
\end{longtable}}
\end{landscape}

\begin{landscape}
\renewcommand{\thetable}{4c}\renewcommand{\theHtable}{S4c}
{\scriptsize\setlength{\tabcolsep}{3pt}
\begin{longtable}{llrrcrrr}
\caption{MIT-BIH sub-annotation stratification of the own-map distance ratio (U1 anchors). S split by PhysioNet symbol: normally conducted atrial/junctional premature beats (A, J, S) versus aberrated atrial premature beats (a); N split by record (records with/without left or right bundle-branch-block beats, L/R) and by beat symbol. Groups need $\geq10$ evaluated beats in a record. Purity: share of true N among the selected U1 anchors (6 seeds pooled).}\label{tab:s4c}\\
\toprule
Set & Group & records & units & ratio median [IQR] & $<$1 (\%) & $>$1 (\%) & U1 purity (\%) \\
\midrule
\endfirsthead
\multicolumn{8}{l}{\textit{(continued)}}\\
\toprule
Set & Group & records & units & ratio median [IQR] & $<$1 (\%) & $>$1 (\%) & U1 purity (\%) \\
\midrule
\endhead
\bottomrule
\endlastfoot
DS1-CV & AAMI S (all) & 7 & 42 & 0.732 [0.638, 0.837] & 92.9 & 7.1 & -- \\
DS1-CV & S: A/J/S (normal-conducted) & 7 & 42 & 0.732 [0.638, 0.836] & 92.9 & 7.1 & -- \\
DS1-CV & S: a (aberrated APC) & 0 & 0 & -- & -- & -- & -- \\
DS2 A & AAMI S (all) & 8 & 48 & 0.755 [0.634, 1.013] & 75.0 & 25.0 & -- \\
DS2 A & S: A/J/S (normal-conducted) & 7 & 42 & 0.732 [0.599, 0.834] & 85.7 & 14.3 & -- \\
DS2 A & S: a (aberrated APC) & 2 & 12 & 1.204 [1.113, 1.306] & 16.7 & 83.3 & -- \\
DS2 B & AAMI S (all) & 8 & 48 & 0.683 [0.537, 1.044] & 72.9 & 27.1 & -- \\
DS2 B & S: A/J/S (normal-conducted) & 7 & 42 & 0.646 [0.529, 0.872] & 88.1 & 11.9 & -- \\
\end{longtable}}
\end{landscape}

\renewcommand{\thetable}{4d}\renewcommand{\theHtable}{S4d}
{\scriptsize\setlength{\tabcolsep}{3pt}
\begin{longtable}{lcrcrr}
\caption{Icentia11k: baseline N--V morphological contrast $d_{\mathrm{LE}}(\bar{C}_N,\bar{C}_V)$ of the record versus the V distance ratio of the own map (1,092 record\,$\times$\,seed units with $\geq10$ evaluated V beats). Spearman $\rho=0.229$ ($p=1.9\times10^{-14}$) at unit level; $\rho=0.213$ ($p=0.0039$) on record means (182 records); Pearson $r=0.214$.}\label{tab:s4d}\\
\toprule
Quartile of $d_{\mathrm{LE}}(\bar{C}_N,\bar{C}_V)$ & range & units & V ratio median [IQR] & $<$1 (\%) & $>$1 (\%) \\
\midrule
\endfirsthead
\multicolumn{6}{l}{\textit{(continued)}}\\
\toprule
Quartile of $d_{\mathrm{LE}}(\bar{C}_N,\bar{C}_V)$ & range & units & V ratio median [IQR] & $<$1 (\%) & $>$1 (\%) \\
\midrule
\endhead
\bottomrule
\endlastfoot
Q1 (smallest) & 0.541--2.126 & 273 & 1.089 [0.895, 1.486] & 35.5 & 61.9 \\
Q2 & 2.127--2.468 & 273 & 1.266 [1.000, 1.622] & 22.0 & 74.0 \\
Q3 & 2.471--2.867 & 273 & 1.353 [1.041, 1.642] & 16.5 & 78.4 \\
Q4 (largest) & 2.870--4.983 & 273 & 1.394 [1.154, 1.677] & 11.4 & 87.2 \\
\end{longtable}}

\clearpage
\section*{Supplementary Table 5: Output-only label-shift comparators, class-weighted BN and prefix class mixture}

\renewcommand{\thetable}{5a}\renewcommand{\theHtable}{S5a}
{\scriptsize\setlength{\tabcolsep}{3pt}
\begin{longtable}{llrrrr}
\caption{Output-only label-shift and non-i.i.d.\ comparators evaluated on the frozen G0 posteriors, alongside the main-text baselines for reference (\%). Prior-EM: EM estimate of the record's class prior from its 5-min prefix (Saerens et al.), evaluation logits corrected by $\log(\pi_t/\pi_s)$; $\pi_s$ uniform (balanced source training) or the source fitting frequencies. LAME: Laplacian-regularized posteriors (Boudiaf et al.) with a $k=5$ nearest-neighbor graph on log-Euclidean vectorized covariances; batch = evaluation segment, prefix + evaluation segment (both transductive) or consecutive 64-beat evaluation batches (online).}\label{tab:s5a}\\
\toprule
Set & Method & $J$ & S-F1 & V-F1 & N-Se \\
\midrule
\endfirsthead
\multicolumn{6}{l}{\textit{(continued)}}\\
\toprule
Set & Method & $J$ & S-F1 & V-F1 & N-Se \\
\midrule
\endhead
\bottomrule
\endlastfoot
DS1-CV & G0 & 39.64 & 18.27 & 61.02 & 75.45 \\
 & Prior-EM (uniform $\pi_s$) & 31.40 & 4.27 & 58.53 & 82.66 \\
 & Prior-EM (source freq.) & 37.67 & 6.03 & 69.32 & 5.29 \\
 & LAME (eval) & 38.31 & 13.74 & 62.87 & 77.89 \\
 & LAME (prefix+eval) & 38.22 & 13.72 & 62.73 & 77.81 \\
 & LAME (64-beat) & 34.23 & 5.82 & 62.65 & 78.41 \\
 & AdaBN & 34.21 & 14.55 & 53.87 & 67.47 \\
 & Tent & 37.49 & 18.10 & 56.88 & 71.47 \\
\end{longtable}}

\begin{landscape}
\renewcommand{\thetable}{5b}\renewcommand{\theHtable}{S5b}
{\scriptsize\setlength{\tabcolsep}{3pt}
\begin{longtable}{llrrrrcrc}
\caption{Prior-EM convergence and minority-class collapse: record\,$\times$\,seed units with a non-empty prefix; \% converged within 100 iterations; median iterations; \% of units whose estimated S or V prior fell below $10^{-6}$, and the count among units whose prefix truly contained that class.}\label{tab:s5b}\\
\toprule
Set & $\pi_s$ & units & conv.\ (\%) & iter. & $\pi_S<10^{-6}$ (\%) & ...with S in prefix & $\pi_V<10^{-6}$ (\%) & ...with V in prefix \\
\midrule
\endfirsthead
\multicolumn{9}{l}{\textit{(continued)}}\\
\toprule
Set & $\pi_s$ & units & conv.\ (\%) & iter. & $\pi_S<10^{-6}$ (\%) & ...with S in prefix & $\pi_V<10^{-6}$ (\%) & ...with V in prefix \\
\midrule
\endhead
\bottomrule
\endlastfoot
DS1-CV & Prior-EM (uniform $\pi_s$) & 126 & 89.7 & 21 & 80.2 & 37/54 & 37.3 & 0/78 \\
DS1-CV & Prior-EM (source freq.) & 126 & 80.2 & 41 & 23.8 & 12/54 & 49.2 & 14/78 \\
DS2 A & Prior-EM (uniform $\pi_s$) & 132 & 94.7 & 16 & 93.9 & 40/48 & 41.7 & 0/66 \\
DS2 A & Prior-EM (source freq.) & 132 & 84.1 & 34 & 18.2 & 5/48 & 50.8 & 8/66 \\
DS2 B & Prior-EM (uniform $\pi_s$) & 132 & 93.2 & 18 & 90.9 & 42/48 & 41.7 & 0/66 \\
DS2 B & Prior-EM (source freq.) & 132 & 81.8 & 42 & 22.7 & 6/48 & 47.7 & 4/66 \\
INCART S44 & Prior-EM (uniform $\pi_s$) & 450 & 87.3 & 23 & 85.8 & 80/108 & 28.0 & 1/318 \\
INCART S44 & Prior-EM (source freq.) & 450 & 83.8 & 40 & 41.1 & 33/108 & 40.0 & 50/318 \\
\end{longtable}}
\end{landscape}

\begin{landscape}
\renewcommand{\thetable}{5c}\renewcommand{\theHtable}{S5c}
{\scriptsize\setlength{\tabcolsep}{3pt}
\begin{longtable}{lllrrrrrrr}
\caption{Class-balanced BN re-estimation (oracle diagnostic): in the single prefix pass of AdaBN, every BatchNorm layer uses weighted batch statistics in which each true class present in the prefix carries equal total weight (prefix labels used; evaluation beats untouched). Records whose prefix holds one class are identical to AdaBN by construction.}\label{tab:s5c}\\
\toprule
Set & subset & records & method & $J$ & V-PPV & V-Se & S-Se & S-PPV & N-Se \\
\midrule
\endfirsthead
\multicolumn{10}{l}{\textit{(continued)}}\\
\toprule
Set & subset & records & method & $J$ & V-PPV & V-Se & S-Se & S-PPV & N-Se \\
\midrule
\endhead
\bottomrule
\endlastfoot
DS2 A & multi-class prefix records & 14 & G0 & 45.02 & 67.45 & 96.66 & 8.51 & 18.09 & 72.89 \\
DS2 A & multi-class prefix records & 14 & GA\_NS & 53.86 & 86.32 & 95.81 & 11.99 & 30.08 & 92.39 \\
DS2 A & multi-class prefix records & 14 & AdaBN & 40.34 & 61.32 & 96.92 & 5.07 & 11.82 & 66.21 \\
DS2 A & multi-class prefix records & 14 & Class-weighted AdaBN & 42.05 & 64.91 & 96.31 & 5.47 & 13.04 & 70.41 \\
DS2 A & all records & 22 & G0 & 47.88 & 64.20 & 96.67 & 18.33 & 21.70 & 80.82 \\
DS2 A & all records & 22 & GA\_NS & 56.04 & 82.91 & 95.82 & 17.99 & 33.75 & 94.14 \\
DS2 A & all records & 22 & AdaBN & 36.71 & 48.68 & 96.93 & 10.93 & 10.12 & 68.98 \\
DS2 A & all records & 22 & Class-weighted AdaBN & 37.79 & 50.36 & 96.32 & 11.26 & 10.81 & 71.58 \\
\end{longtable}}
\end{landscape}

\begin{landscape}
\renewcommand{\thetable}{5d}\renewcommand{\theHtable}{S5d}
{\scriptsize\setlength{\tabcolsep}{3pt}
\begin{longtable}{llrrrrrrrrr}
\caption{Pooled $J$ (\%) by the true N share of the record's eligible prefix beats (half-open bins; records with an empty prefix listed separately, where every method equals G0 under the empty-prefix fallback rule). S, V: true evaluated beats per seed in the stratum.}\label{tab:s5d}\\
\toprule
Set & prefix N & records & S & V & G0 & AdaBN & Tent & T3A & RPA-Recenter & GA\_NS \\
\midrule
\endfirsthead
\multicolumn{11}{l}{\textit{(continued)}}\\
\toprule
Set & prefix N & records & S & V & G0 & AdaBN & Tent & T3A & RPA-Recenter & GA\_NS \\
\midrule
\endhead
\bottomrule
\endlastfoot
Icentia11k all & $<$80\% & 16 & 6,697 & 9,717 & 61.04 & 50.84 & 52.67 & 54.57 & 61.77 & 61.57 \\
 & 80-90\% & 21 & 4,694 & 5,494 & 59.98 & 53.00 & 54.19 & 51.89 & 63.96 & 68.99 \\
 & 90-95\% & 26 & 3,497 & 2,206 & 39.05 & 35.60 & 38.19 & 36.73 & 46.19 & 56.55 \\
 & 95-99\% & 103 & 3,716 & 5,385 & 49.90 & 39.89 & 47.01 & 41.61 & 58.07 & 66.43 \\
 & $\geq$99\% & 772 & 3,957 & 3,074 & 10.78 & 7.03 & 9.40 & 5.60 & 15.50 & 21.85 \\
 & empty prefix & 35 & 515 & 1,412 & 51.83 & 51.83 & 51.83 & 51.83 & 51.83 & 51.83 \\
\midrule
DS1-CV & $<$80\% & 3 & 108 & 1,297 & 33.13 & 34.52 & 34.47 & 43.75 & 49.67 & 47.04 \\
 & 80-90\% & 2 & 0 & 833 & 35.84 & 35.48 & 34.77 & 42.55 & 43.33 & 36.53 \\
\end{longtable}}
\end{landscape}

\clearpage

\begin{landscape}
\section*{Supplementary Table 6: Unclassified prefix beats, fiducial jitter and annotation-free detection}
\renewcommand{\thetable}{6a}\renewcommand{\theHtable}{S6a}
{\scriptsize\setlength{\tabcolsep}{3pt}
\begin{longtable}{lrrrrrrrrr}
\caption{Icentia11k: unclassified (Q) beats admitted to the 5-min prefix without pre-exclusion (evaluation segment unchanged). Counts per seed (6-seed mean). Q removed at stage 1 (RR filter: $\Delta_i/m_i\geq0.9$ and $p(\mathrm{N})\geq0.5$) and stage 2 (80\% LE core), as \% of all prefix Q beats.}\label{tab:s6a}\\
\toprule
Prefix & beats & Q & Q rm.\ RR (\%) & Q rm.\ core (\%) & Q kept (\%) & U1 anchors & anchor N (\%) & anchor Q (\%) & non-empty prefix \\
\midrule
\endfirsthead
\multicolumn{10}{l}{\textit{(continued)}}\\
\toprule
Prefix & beats & Q & Q rm.\ RR (\%) & Q rm.\ core (\%) & Q kept (\%) & U1 anchors & anchor N (\%) & anchor Q (\%) & non-empty prefix \\
\midrule
\endhead
\bottomrule
\endlastfoot
Q excluded (frozen) & 272,509 & 0 & -- & -- & -- & 161,079 & 99.82 & 0.00 & 938 \\
Q admitted & 347,354 & 74,845 & 47.5 & 17.1 & 35.4 & 191,092 & 86.00 & 13.85 & 969 \\
\end{longtable}}
\end{landscape}

\renewcommand{\thetable}{6b}\renewcommand{\theHtable}{S6b}
{\scriptsize\setlength{\tabcolsep}{3pt}
\begin{longtable}{lrlrrr}
\caption{Icentia11k $J$ (\%) with Q beats excluded (primary protocol) or admitted in the prefix.}\label{tab:s6b}\\
\toprule
Cohort & records & method & $J$ (Q excluded) & $J$ (Q admitted) & $\Delta J$ (pp) \\
\midrule
\endfirsthead
\multicolumn{6}{l}{\textit{(continued)}}\\
\toprule
Cohort & records & method & $J$ (Q excluded) & $J$ (Q admitted) & $\Delta J$ (pp) \\
\midrule
\endhead
\bottomrule
\endlastfoot
all-comers & 973 & G0 & 34.54 & 34.54 & $+$0.00 \\
all-comers & 973 & GA\_NS & 48.66 & 48.67 & $+$0.01 \\
all-comers & 973 & RPA-Recenter & 42.75 & 43.69 & $+$0.94 \\
non-AF & 919 & G0 & 43.87 & 43.87 & $+$0.00 \\
non-AF & 919 & GA\_NS & 59.81 & 60.02 & $+$0.21 \\
non-AF & 919 & RPA-Recenter & 54.82 & 56.02 & $+$1.20 \\
35 empty-prefix records & 35 & G0 & 51.83 & 51.83 & $+$0.00 \\
35 empty-prefix records & 35 & GA\_NS & 51.83 & 51.08 & $-$0.74 \\
35 empty-prefix records & 35 & RPA-Recenter & 51.83 & 51.55 & $-$0.28 \\
\end{longtable}}

\renewcommand{\thetable}{6c}\renewcommand{\theHtable}{S6c}
{\scriptsize\setlength{\tabcolsep}{3pt}
\begin{longtable}{llrrrrrrr}
\caption{Fiducial jitter: $J$ with annotation fiducials and $J$ retention (\% of the annotation $J$) when every realigned fiducial is displaced by uniform ($\pm a$\,ms) or Gaussian ($\sigma=a$\,ms) jitter; crop positions and all RR features move; labels and evaluated beat sets unchanged.}\label{tab:s6c}\\
\toprule
Set & method & $J$ (annot.) & U$\pm$10 & U$\pm$25 & U$\pm$50 & G10 & G25 & G50 \\
\midrule
\endfirsthead
\multicolumn{9}{l}{\textit{(continued)}}\\
\toprule
Set & method & $J$ (annot.) & U$\pm$10 & U$\pm$25 & U$\pm$50 & G10 & G25 & G50 \\
\midrule
\endhead
\bottomrule
\endlastfoot
DS2 A & G0 & 47.88 & 98.0 & 96.5 & 91.6 & 97.9 & 92.9 & 86.3 \\
 & RPA-Recenter & 50.34 & 99.5 & 99.6 & 97.0 & 99.6 & 98.3 & 93.4 \\
 & GA\_NS & 56.04 & 98.5 & 98.6 & 96.4 & 98.1 & 97.4 & 92.4 \\
\midrule
DS2 B & G0 & 44.07 & 96.7 & 95.2 & 90.0 & 96.0 & 91.8 & 83.0 \\
 & RPA-Recenter & 46.56 & 99.5 & 98.7 & 94.3 & 99.0 & 96.0 & 89.2 \\
 & GA\_NS & 55.61 & 99.0 & 98.4 & 94.0 & 99.0 & 95.8 & 88.3 \\
\midrule
INCART S44 & G0 & 67.94 & 99.1 & 96.4 & 84.8 & 98.0 & 88.1 & 70.8 \\
 & RPA-Recenter & 67.62 & 99.8 & 98.2 & 88.0 & 98.9 & 91.0 & 72.7 \\
\end{longtable}}

\renewcommand{\thetable}{6d}\renewcommand{\theHtable}{S6d}
{\scriptsize\setlength{\tabcolsep}{2pt}
\setlength{\LTpost}{2pt}
\begin{longtable}{lrrrrrrrr}
\caption{Annotation-free pipeline with the open-source \texttt{xqrs} detector (WFDB Python package, v4.3.1) on the frozen 360\,Hz signal; peaks realigned with the frozen fiducial rule. Matching to reference beats: one-to-one, greedy by offset, $\pm150$\,ms (EC57). Detection Se and PPV over the evaluation segment.}\label{tab:s6d}\\
\toprule
Set & records & ref.\ beats & detections & missed & unmatched & excluded (Q/non-eval.) & Se (\%) & PPV (\%) \\
\midrule
\endfirsthead
\multicolumn{9}{l}{\textit{(continued)}}\\
\toprule
Set & records & ref.\ beats & detections & missed & unmatched & excluded (Q/non-eval.) & Se (\%) & PPV (\%) \\
\midrule
\endhead
\bottomrule
\endlastfoot
DS2 A & 22 & 41,478 & 41,401 & 133 & 49 & 7 & 99.68 & 99.88 \\
DS2 B & 22 & 41,478 & 41,401 & 133 & 49 & 7 & 99.68 & 99.88 \\
INCART S44 & 75 & 147,464 & 145,814 & 2,648 & 971 & 27 & 98.20 & 99.33 \\
INCART DS1-21 & 75 & 147,464 & 145,814 & 2,648 & 971 & 27 & 98.20 & 99.33 \\
SVDB S44 & 78 & 153,652 & 153,151 & 900 & 342 & 57 & 99.41 & 99.78 \\
SVDB DS1-21 & 78 & 153,652 & 153,151 & 900 & 342 & 57 & 99.41 & 99.78 \\
Icentia11k all$^{\dagger}$ & 973 & 3,615,388 & 4,428,963 & 211,844 & 159,210 & 866,209 & 94.14 & 96.41 \\
\end{longtable}
\par\noindent{\footnotesize $^{\dagger}$Single-channel chest patch; xqrs was applied without patch-specific tuning, and its detection errors (Se 94.1\%, PPV 96.4\%) propagate into the classification metrics of Supplementary Table~6e.\par}}

\begin{landscape}
\renewcommand{\thetable}{6e}\renewcommand{\theHtable}{S6e}
{\scriptsize\setlength{\tabcolsep}{1.8pt}
\setlength{\LTpost}{2pt}
\begin{longtable}{llrrrrrrrrrr}
\caption{$J$ (\%) of the annotation-driven and the detector-driven (\texttt{xqrs}) pipelines and their differences (pp). Detector-driven metrics: an evaluated reference beat without a matched detection counts as missed (false negative of its class); an unmatched detection labeled S or V counts as a false positive of that class; detections matched to Q or non-evaluated reference beats are excluded.}\label{tab:s6e}\\
\toprule
Set & method & $J$ annot. & $J$ det. & det.$-$annot. & vs G0 (annot.) & vs G0 (det.) & S-Se & S-PPV & V-Se & V-PPV & N-Se \\
\midrule
\endfirsthead
\multicolumn{12}{l}{\textit{(continued)}}\\
\toprule
Set & method & $J$ annot. & $J$ det. & det.$-$annot. & vs G0 (annot.) & vs G0 (det.) & S-Se & S-PPV & V-Se & V-PPV & N-Se \\
\midrule
\endhead
\bottomrule
\endlastfoot
DS2 A & G0 & 47.88 & 45.92 & $-$1.96 & $+$0.00 & $+$0.00 & 17.1 & 19.9 & 93.1 & 62.3 & 80.5 \\
 & AdaBN & 36.71 & 34.83 & $-$1.88 & $-$11.17 & $-$11.09 & 11.8 & 9.7 & 93.3 & 45.5 & 67.8 \\
 & Tent & 39.47 & 37.37 & $-$2.10 & $-$8.41 & $-$8.55 & 10.6 & 12.9 & 92.0 & 51.4 & 73.9 \\
 & RPA-Recenter & 50.34 & 48.72 & $-$1.62 & $+$2.46 & $+$2.80 & 17.3 & 23.8 & 78.0 & 77.2 & 95.8 \\
 & GA\_NS & 56.04 & 53.92 & $-$2.12 & $+$8.17 & $+$8.00 & 17.3 & 31.3 & 92.2 & 80.5 & 94.0 \\
\midrule
DS2 B & G0 & 44.07 & 42.67 & $-$1.40 & $+$0.00 & $+$0.00 & 19.6 & 17.8 & 93.1 & 52.4 & 81.4 \\
 & AdaBN & 30.98 & 30.12 & $-$0.86 & $-$13.09 & $-$12.55 & 7.6 & 6.9 & 93.6 & 38.0 & 66.0 \\
 & Tent & 34.58 & 33.66 & $-$0.92 & $-$9.49 & $-$9.01 & 9.1 & 9.9 & 91.7 & 43.9 & 72.1 \\
\end{longtable}
\par\noindent{\footnotesize $^{\dagger}$See the footnote of Supplementary Table~6d.\par}}
\end{landscape}

\end{document}